\documentclass[journal]{IEEEtranTIE}
\usepackage{graphicx}
\usepackage{cite}
\usepackage{picinpar}
\usepackage{amsmath}
\usepackage{url}
\usepackage{flushend}
\usepackage[latin1]{inputenc}
\usepackage{colortbl}
\usepackage{soul}
\usepackage{multirow}
\usepackage{pifont}
\usepackage{color}
\usepackage{alltt}
\usepackage[hidelinks]{hyperref}
\usepackage{enumerate}
\usepackage{siunitx}
\usepackage{epstopdf}
\usepackage{pbox}
\usepackage{amsfonts}
\usepackage{bm}
\usepackage{empheq}
\usepackage{amsthm}
\usepackage[caption=false, font=footnotesize]{subfig}
\usepackage{algorithm2e}
\usepackage{booktabs}
\newtheoremstyle{assumptionstyle}
{3pt}
{3pt}
{\normalfont}
{\parindent}
{\itshape}
{.}
{.5em}
{}
\theoremstyle{assumptionstyle}

\usepackage{mdframed}
\usepackage{xcolor}

\newmdenv[
  backgroundcolor=yellow,
  linecolor=white,
  innertopmargin=0pt,
  innerbottommargin=0pt,
  innerleftmargin=0pt,
  innerrightmargin=0pt,
  skipabove=0pt,
  skipbelow=0pt
]{highlight}

\begin{document}
\title{Learning-Enhanced Observer for Sensorless Task-Null-Space Compliant Control}
\author{
	\vskip 1em
	Nuo~Chen, \emph{Student Member,~IEEE},
	Ya-Jun~Pan, \emph{Senior~Member,~IEEE},
    and~Onur~Toker, \emph{Member,~IEEE}

	\thanks{
		This work was supported by the Natural Sciences and Engineering Research Council (NSERC), FulBright Canada Foundation, and the Government of Nova Scotia, Canada. (Corresponding author: Ya-Jun Pan)
  
		Nuo Chen and Ya-Jun Pan are with the Advanced Control and Mechatronics Lab, Department of Mechanical Engineering, Dalhousie University, Halifax, B3H 4R2, Canada (e-mails: nuo.chen@dal.ca, yajun.pan@dal.ca).

        Onur Toker is with the Department of Computer Engineering, Florida Polytechnic University, Lakeland, FL 33805, United States (e-mail: otoker@floridapoly.edu).
	}
}
\maketitle
	
\begin{abstract}
This paper presents a novel framework integrating a robust wrench observer with a hybrid pose and force controller for physical Human-Robot Interaction (pHRI). In pHRI tasks, robots are expected to perform task-space motions robustly while remaining compliance with human disturbance, which proposes distinct control objectives. Most existing approaches focus mainly on either robustness or compliance and rely on force sensors for real-time tracking. To address these limitations, we propose a novel framework with the following key contributions. Firstly, a neural network (NN) is trained to improve the observer performance by compensating for system friction and unmodeled dynamics. Secondly, a task-null-space decomposition is applied to separate the joint velocity and reduce the effect of the human disturbance on the observer. Finally, a hybrid pose and force controller is introduced to achieve the end-effector pose and force tracking. Experimental results demonstrate that the proposed framework provides robust and accurate wrench estimation while maintaining reliable pose and force tracking performance. The proposed approach offers an effective solution and framework for pHRI without the need of force/torque sensor measurements which can be applied to many robotic manipulator platforms which do not have force/torque sensors. 
\end{abstract}

\begin{IEEEkeywords}
Physical human-robot interaction, Neural network, Observer, Decomposition, Robotic manipulators.
\end{IEEEkeywords}

\markboth{IEEE TRANSACTIONS ON INDUSTRIAL ELECTRONICS}%
{}
\definecolor{limegreen}{rgb}{0.2, 0.8, 0.2}
\definecolor{forestgreen}{rgb}{0.13, 0.55, 0.13}
\definecolor{greenhtml}{rgb}{0.0, 0.5, 0.0}

\section{Introduction}

\IEEEPARstart{P}{hysical} human-robot interaction (pHRI) has received increasing attention due to the need of humans and robots to share the same workspace. Compared with traditional robotic tasks, pHRI requires the robot to achieve two different objectives at the same time. On one hand, the robot should accurately complete its task, such as position, orientation, and force tracking. On the other hand, it should remain compliant when unexpected human interaction occurs to ensure safety. Therefore, an effective pHRI framework should provide accurate task execution, reliable interaction state estimation, and compliant response to human disturbances.
\begin{figure*}[t]
    \centering
    \includegraphics[width=1\linewidth]{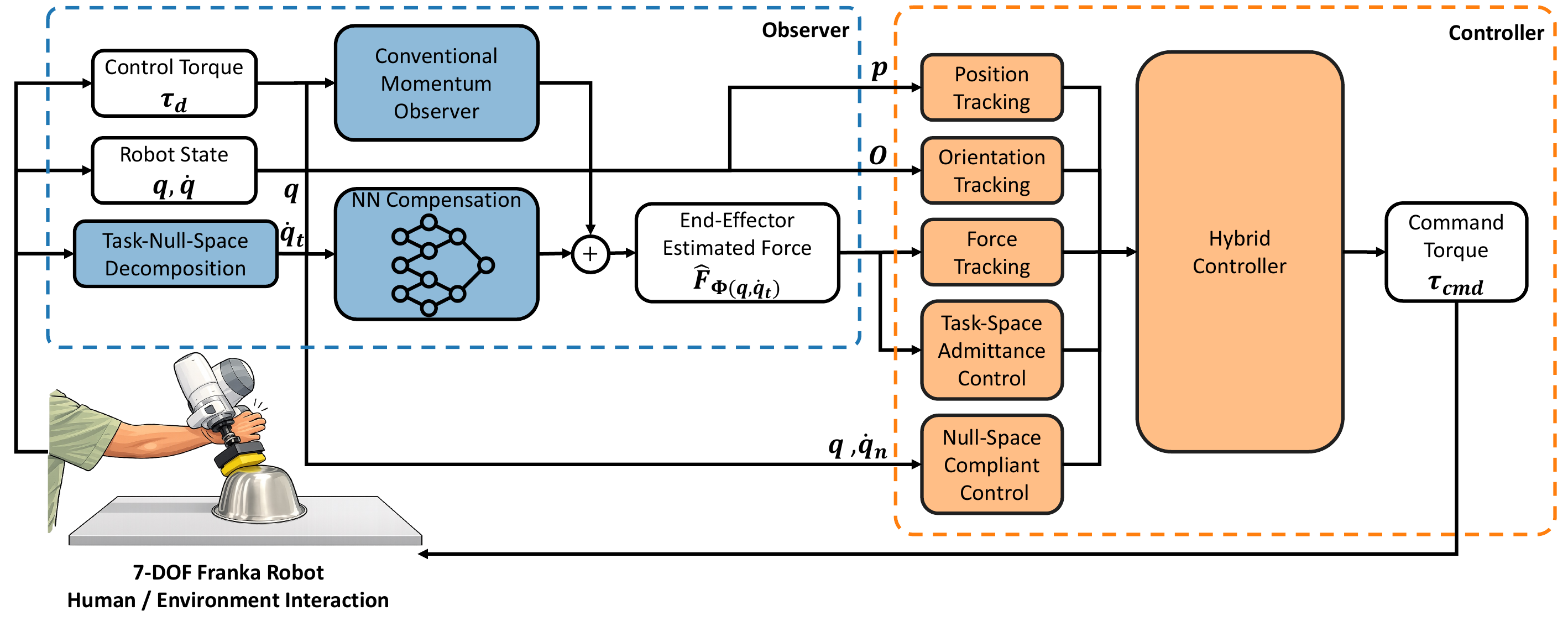}
    \caption{Overview of the proposed framework. The observer estimates the end-effector force using a momentum observer and NN compensation. A task-null-space decomposition separates the joint velocity $\dot{\bm{q}}$ into the task-space velocity $\dot{\bm{q}}_t$ and null-space velocity $\dot{\bm{q}}_n$. The task-space velocity $\dot{\bm{q}}_t$ is sent to the NN to reduce the effect of human disturbance on force estimation, while $\dot{\bm{q}}_n$ is used for null-space compliant control. The proposed hybrid pose and force controller provides reliable position, orientation, and force tracking while maintaining compliance in the null-space.}
    \label{fig:Diagram}
\end{figure*}

Force control is necessary for robotic tasks which involve physical contact with humans or the environment. In literature, commonly used force control approaches include the direct force control~\cite{Chen_2026_ISIE}, hybrid position and force control~\cite{fang_2026_Arxiv, Han_2021_TII}, impedance control~\cite{Sharifi_2022_TCST}, and admittance control~\cite{Li_2026_CEP, Wan_2023_AIM}. The hybrid position and force control usually regulates motion and interaction force in different task directions, while the impedance and admittance control regulate the relationship between the motion and force. Some recent studies have combined these control strategies to achieve both accurate force tracking and compliant motion. Hybrid force-impedance controllers were developed to achieve force tracking during fast end-effector motions~\cite{Iskandar_2023_RAL} and large force pulling~\cite{Xu_2025_Arxiv}. In \cite{Haddadin_2024_IJRR}, a framework to combine force regulation and compliant behavior within a passivity formulation is presented. In \cite{Ye_2024_Robotica},  an impedance-admittance framework to improve compliance and adaptability in changing conditions is proposed. These studies have shown the effectiveness of hybrid control in different interaction tasks. However, the force regulation and compliant control often require a reliable interaction force measurements from force and torque sensors.

Access to the interaction force is an important part of pHRI. Force and torque sensors can provide direct measurements of external wrench~\cite{Huo_2025_Actuators, Al_2021_RCIM}, but they often increase the hardware cost and may not be available in all robotic systems. Force/torque measurements often are noisy and need signal processing before applied to the controller. Observers provide an alternative solution by estimating end-effector wrench from robot dynamics and available joint motion~\cite{Cao_2021_RCIM}. Among them, momentum observers are commonly used because they do not require noisy acceleration~\cite{Park_2022_TMECH, Haddadin_2017_TRO}. However, the estimation accuracy may be affected by joint friction, unmodeled dynamics, and control errors~\cite{Kommuri_2022_TMech}. To address these limitations, recent studies have applied adaptive~\cite{Tsai_2026_RCIM} and learning-based compensation~\cite{Liu_2021_RCIM} to improve the estimation performance. Moreover, for redundant manipulators, external disturbances may generate joint motion that does not contribute to the task-space motion~\cite{Yang_2026_TMRB}. Such motion can appear in the null-space while the end-effector task remains almost unchanged~\cite{Yang_2025_RAL}. However, conventional momentum observers are usually constructed using the overall joint motion, meaning that the human null-space motion may be incorrectly reflected in the task-space estimation. This issue is especially important in pHRI, where human applied disturbance cannot be fully rejected but should be handled differently from the task-related motion.

In pHRI, robustness and compliance are often treated as two independent objectives. Robust controllers are designed to reject external disturbances and maintain accurate task execution, while compliant controllers allow the robot to respond compliantly to human and/or the environment interaction. Most existing approaches mainly focus on designing different controllers and switching between them according to different states~\cite{Bai_2021_TRO, Xu_2021_RCIM, Formenti_2022_CES, Zhou_2024_TMECH}, making it difficult to achieve both tracking and disturbance rejection objectives at the same time. Moreover, many studies on the pose and force control are conducted on 2D-flat surfaces~\cite{Yu_2024_TIE, Tseng_2025_IJFS, Hung_2023_AS}, limiting their applicability to real-world tasks. Motivated by these challenges, we aim to achieve the robust end-effector task performance while allowing a compliant motion under human applied disturbance. For a redundant manipulator, human interaction can be accommodated through null-space motion without changing the desired end-effector behavior. The overview of the proposed framework is illustrated in Fig.~\ref{fig:Diagram}. The proposed framework consists of an observer module and a hybrid controller for simultaneous task execution and compliant pHRI. The observer combines a conventional momentum observer with a task-null-space decomposition and neural-network (NN) compensation to estimate the end-effector interaction force while reducing the influence of null-space human disturbances. Based on the estimated force, the controller integrates position, orientation, and force tracking with task-space admittance and null-space compliant control to generate the commanded torque, enabling a desired robust task-space performance together with a compliant interaction with human.

The main contributions of this work are summarized as follows:
\begin{enumerate}
  \item A new NN-enhanced momentum observer with task-null-space interaction decomposition is proposed for sensorless external force estimation. The NN compensates for observation errors caused by joint friction and model uncertainties. The proposed decomposition separates the estimated task-space force from null-space human interactions. Compared with the learning-based observer in~\cite{Chen_2026_ISIE} and~\cite{Liu_2021_RCIM}, this design eliminates the influence of null-space disturbances on the NN-enhanced observer.
  
  \item A hybrid pose and force controller is developed by integrating the Cartesian position and orientation tracking, force regulation, task-space admittance control, and null-space compliant control. Unlike the 2D-flat surface tasks in~\cite{Yu_2024_TIE} and~\cite{Tseng_2025_IJFS}, the proposed controller enables the simultaneous position, orientation, and force tracking on a 3D-curved surface. Meanwhile, it achieves a compliant behavior to ensure both task robustness and pHRI safety at the same time instead of the need of mode switching as presented in~\cite{Bai_2021_TRO} and~\cite{Zhou_2024_TMECH}, which may cause instability.

  \item The proposed observer-controller framework is evaluated on a 7-degree-of-freedom (DoF) Franka Emika Panda robot. Experimental results under different trajectories and interaction scenarios demonstrate an improved force estimation performance, accurate pose and force tracking, and effective compliant responses in null-space.

\end{enumerate}

  In summary, the proposed approach offers a new and feasible solution and framework for pHRI without the need of force/torque sensor measurements. As a result, the new framework can be applied to many robotic manipulator platforms which often do not have force/torque sensors. One example is that for underwater robotic manipulators, it is challenging to install force/torque sensors at the end-effector in the underwater environment. 

\section{System Description}
\subsection{Robot Manipulator Dynamics}
\begin{figure}[b]
    \centering
    \subfloat[]{
        \includegraphics[width=0.46\linewidth]{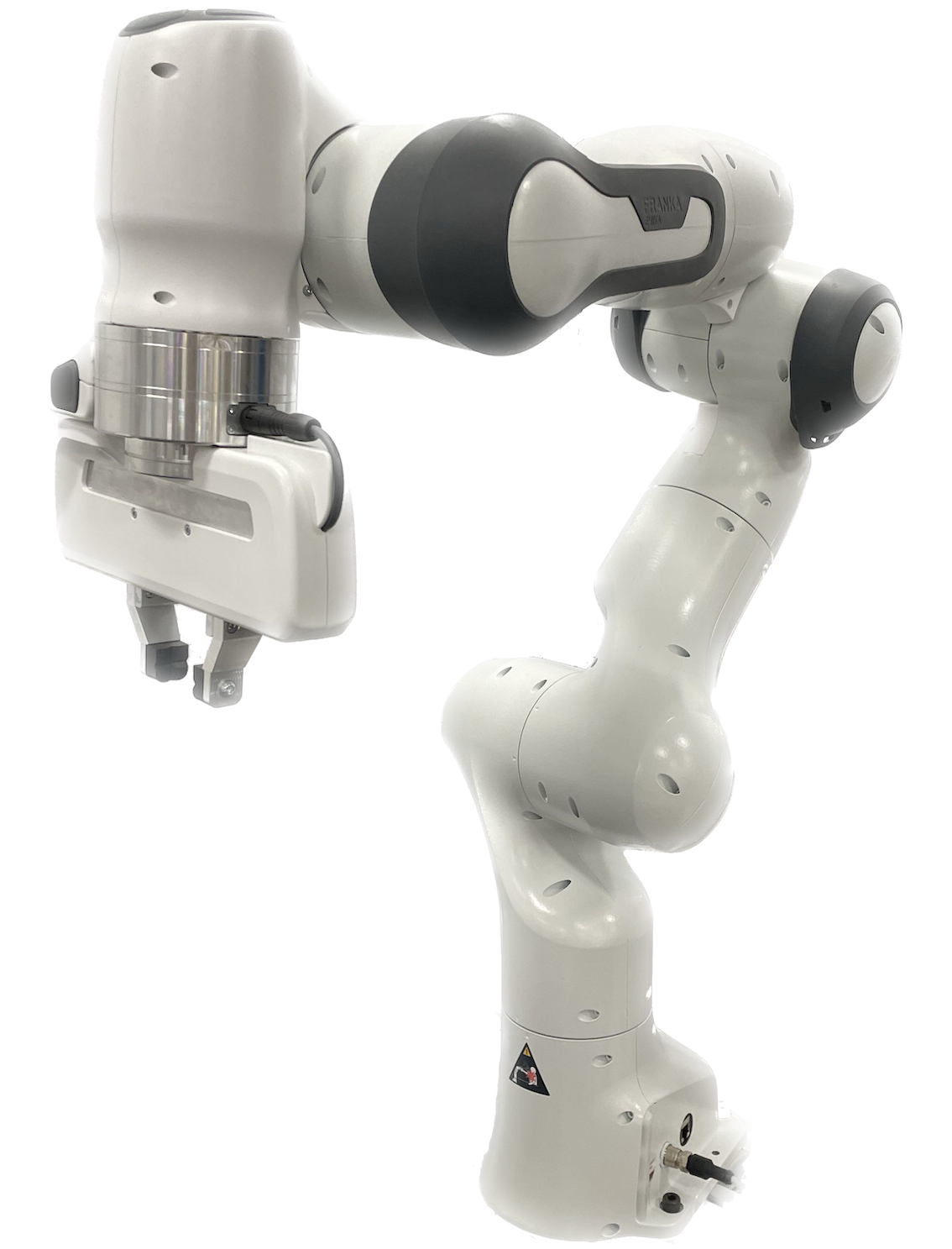}}
    \subfloat[]{
        \includegraphics[width=0.54\linewidth]{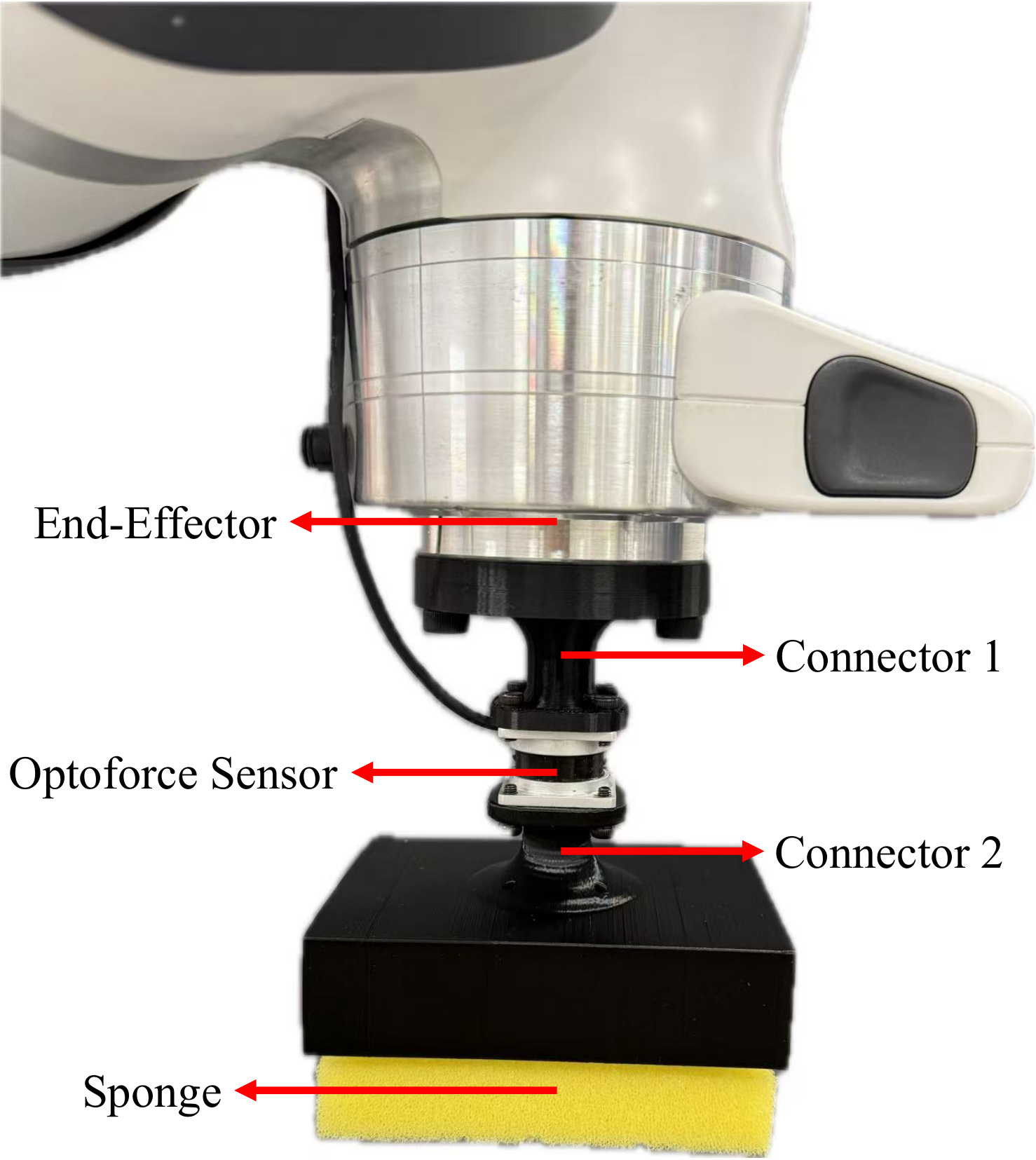}}
    \caption{Hardware Setup: (a) A 7-DOF Franka Emika robotic manipulator in the Advanced Control and Mechatronics Laboratory, Dalhousie University. (b) An end-effector installed with force sensor for data collection and validation.}
    \label{fig:Hardware} 
\end{figure}

Consider an n-DOF robot manipulator, the dynamics of the manipulator in joint-space are expressed as:
\begin{equation}
    \mathcal{M}(\bm{q})\bm{\ddot{q}}+\mathcal{C}(\bm{q},\bm{\dot{q}})\bm{\dot{q}}+\mathcal{G}(\bm{q})=\bm{\tau}_\mathrm{cmd}+\bm{\tau}_d,\label{dynamic_js}
\end{equation}
where $\bm{q}$, $\bm{\dot{q}}$, and $\bm{\ddot{q}} \in \mathbb{R}^{n}$ are the joint position, velocity, and acceleration vectors, respectively. $\mathcal{M}(\bm{q}) \in \mathbb{R}^{n\times n}$ denotes the inertia matrix, $\mathcal{C}(\bm{q}, \bm{\dot{q}}) \in \mathbb{R}^{n\times n}$ represents the Coriolis and centripetal matrix, and $\mathcal{G}(\bm{q}) \in \mathbb{R}^{n}$ is the gravity vector. $\bm{\tau}_\mathrm{cmd} \in \mathbb{R}^{n}$ denotes the commanded torque input vector, and $\bm{\tau}_d = \bm{\tau}_f + \bm{\tau}_u + \bm{\tau}_e \in \mathbb{R}^{n}$ represents the lumped disturbance torque, which is the summation of the internal friction $\bm{\tau}_f$, unmodeled dynamics $\bm{\tau}_u$, and external torque exerted on the manipulator $\bm{\tau}_e$. As shown in Fig.~\ref{fig:Hardware}(a), the manipulator used in this study is a 7-DOF Franka Emika robot with $3$ Kg payload. It provides a real-time joint torque control interface with a 1 kHz sampling rate.

In Cartesian space, the dynamics of the manipulator can be represented as
\begin{equation} 
    \mathcal{\bar{M}}(\bm{q})\bm{\ddot{x}}+\mathcal{\bar{C}}(\bm{q},\bm{\dot{q}})\bm{\dot{x}}+\mathcal{\bar{G}}(\bm{q})=\bm{u} + \bm{u}_d,\label{dynamic_cs} 
\end{equation}
where the end-effector pose is denoted as 
$\bm{x} = \begin{bmatrix}
\bm{p}^T, & \bm{\mathcal{Q}}^T
\end{bmatrix}^T\in\mathbb{R}^{7}$
, with $\bm{p} \in\mathbb{R}^{3}$ is the translational position and
$\bm{\mathcal{Q}}\in\mathbb{R}^4$
is the orientation in the form of a unit quaternion. The end-effector velocity and acceleration are denoted as $\dot{\bm{x}} = 
\begin{bmatrix}
    \dot{\bm{p}}^T, & \bm{\omega}^T
\end{bmatrix}^T \in \mathbb{R}^{6}, \
\ddot{\bm{x}} =
\begin{bmatrix}
    \ddot{\bm{p}}^T, & \dot{\bm{\omega}}^T
\end{bmatrix}^T \in \mathbb{R}^{6}$, with $\dot{\bm{p}}, \ddot{\bm{p}} \in \mathbb{R}^3$ are the translational velocity and acceleration, $\bm{\omega}, \dot{\bm{\omega}} \in \mathbb{R}^3$ are the angular velocity and acceleration. $\bm{u}\in\mathbb{R}^{6}$ is the control input and $\bm{u}_d \in\mathbb{R}^{6}$ represents the equivalent task-space disturbance.  

The Cartesian parametric matrices in \eqref{dynamic_cs} are related to the joint-space matrices in \eqref{dynamic_js} as
\begin{align*}
    \bar{\mathcal{M}}(\bm{q}) &= J(\bm{q})^{-T} \mathcal{M}(\bm{q}) J(\bm{q})^{-1}, \\
    \bar{\mathcal{C}}(\bm{q},\bm{\dot{q}}) &= J(\bm{q})^{-T} \Big( \mathcal{C}(\bm{q},\bm{\dot{q}}) 
     - \mathcal{M}(\bm{q}) J(\bm{q})^{-1} \dot{J}(\bm{q}) \Big) J(\bm{q})^{-1},\\
    \bar{\mathcal{G}}(\bm{q}) &= J(\bm{q})^{-T} \mathcal{G}(\bm{q}),\hspace{1em}\\
    \bm{u} &= J(\bm{q})^{-T}\bm{\tau},
\end{align*}
where $J(\bm{q})\in \mathbb{R}^{6\times n}$ is the Jacobian matrix.

\textit{Property 1}: The inertia matrix $\mathcal{M}(\bm{q})$ is symmetric positive definite, and the matrix $\dot{\mathcal{M}}(\bm{q}) - 2\mathcal{C}(\bm{q}, \bm{\dot{q}})$ is skew-symmetric and it follows that $\dot{\mathcal{M}}(\bm{q}) = \mathcal{C}(\bm{q}, \bm{\dot{q}}) + \mathcal{C}^{T}(\bm{q}, \bm{\dot{q}})$.

\subsection{End-Effector Configuration}
As shown in Fig.~\ref{fig:Hardware}(b), the Franka Emika gripper is replaced by two connectors, one 3D force sensor, and one sponge to perform the wiping task. The end-effector is mounted with a fixed offset $l_s=0.09~\mathrm{m}$ along the normal direction. The optoforce three-axis optical force sensor (OMD-20-FG-100N) is employed for observer data collection and result validation. The sensor operates based on an infrared LED and photodiode array to detect strain applied to the outer rubber surface \cite{Wan_2020_CSME}. The sensor calibration procedure is described as in \cite{Chen_2026_ISIE}.

\section{Observer Design}
\subsection{Momentum Observer}
To enable sensorless force estimation in the tracking process, a force observer is designed to replace the force sensor. The generalized momentum observer was originally proposed to eliminate the need for joint acceleration estimates, which are usually noisy in robotic systems. Moreover, the momentum observer avoids the inversion of the robot inertia matrix $\mathcal{M}(\bm{q})$.

The generalized momentum $\bm{p} \in \mathbb{R}^{7}$ of the manipulator is defined as
\begin{equation}
    \bm{p} = \mathcal{M}(\bm{q})\dot{\bm{q}}.
    \label{eq:momentum}
\end{equation}

The derivative of the momentum in \eqref{eq:momentum} is derived as
\begin{equation}
    \dot{\bm{p}}
    =
    \mathcal{M}(\bm{q})\ddot{\bm{q}}
    +
    \dot{\mathcal{M}}(\bm{q})\dot{\bm{q}}.
    \label{eq:momentum_dot}
\end{equation}

Combining the joint-space dynamics in \eqref{dynamic_js} with
\eqref{eq:momentum_dot} and applying \textit{Property~1}, the momentum dynamics can be rewritten as
\begin{equation}
    \dot{\bm{p}}
    =
    \bm{\tau}_\mathrm{cmd}
    +
    \mathcal{C}^{T}(\bm{q},\dot{\bm{q}})\dot{\bm{q}}
    -
    \mathcal{G}(\bm{q})
    +
    \bm{\tau}_d.
    \label{eq:momentum_simplified}
\end{equation}

A residual signal $\bm{r} \in \mathbb{R}^{7}$ is introduced to act as an estimated disturbance torque $\hat{\bm{\tau}}_d$ for the disturbance torque $\bm{\tau}_d$, defining as
\begin{equation}
    \hat{\bm{\tau}}_d
    =
    \bm{r}
    =
    K_{O}
    \left(
    \bm{p}-\hat{\bm{p}}
    \right),
    \label{eq:residual}
\end{equation}
where $K_{O} = \text{diag}\{k_{O,i}\} > 0 \in \mathbb{R}^{7 \times 7}$ is the diagonal gain matrix for the observer. In this work, $k_{O}=10$. According to the momentum dynamics \eqref{eq:momentum_simplified} and the residual definition \eqref{eq:residual}, the observer update law is defined as
\begin{align}
    \dot{\hat{\bm{p}}} &= \bm{\tau}_\mathrm{cmd} + \mathcal{C}^{T}(\bm{q},\dot{\bm{q}})\dot{\bm{q}} - \mathcal{G}(\bm{q}) + \bm{r},\\
    \dot{\bm{r}} &= K_{O} (\dot{\bm{p}}-\dot{\hat{\bm{p}}}).
    \label{eq:update_law}
\end{align}

The momentum observer output $\bm{r}(t) \in \mathbb{R}^{7}$ is expressed as
\begin{align}
    \bm{r}
    = K_{O} \Bigl[
    &\bm{p}(t) - \bm{p}(0)\notag\\
    &
    - \int_0^t
    \left(
    \bm{\tau}_\mathrm{cmd}
    + \mathcal{C}^{T}(\bm{q},\dot{\bm{q}})\dot{\bm{q}}
    - \mathcal{G}(\bm{q})
    + \bm{r}
    \right)
    \, ds
    \Bigr].
    \label{eq:observer_output}
\end{align}

\begin{figure}[ht]
    \centering
    \subfloat[]{
        \includegraphics[width=0.5\linewidth]{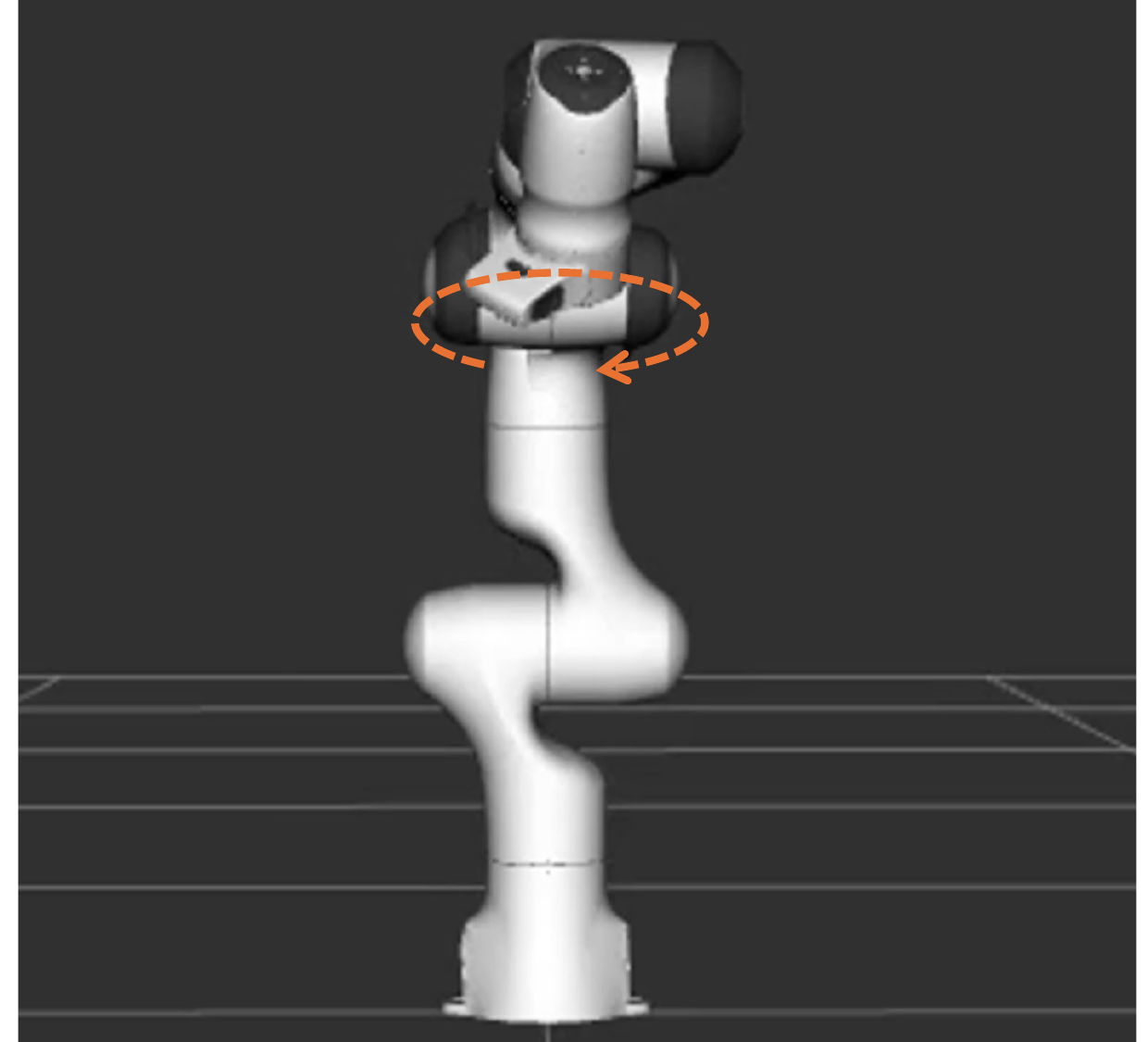}}
    \subfloat[]{
        \includegraphics[width=0.43\linewidth]{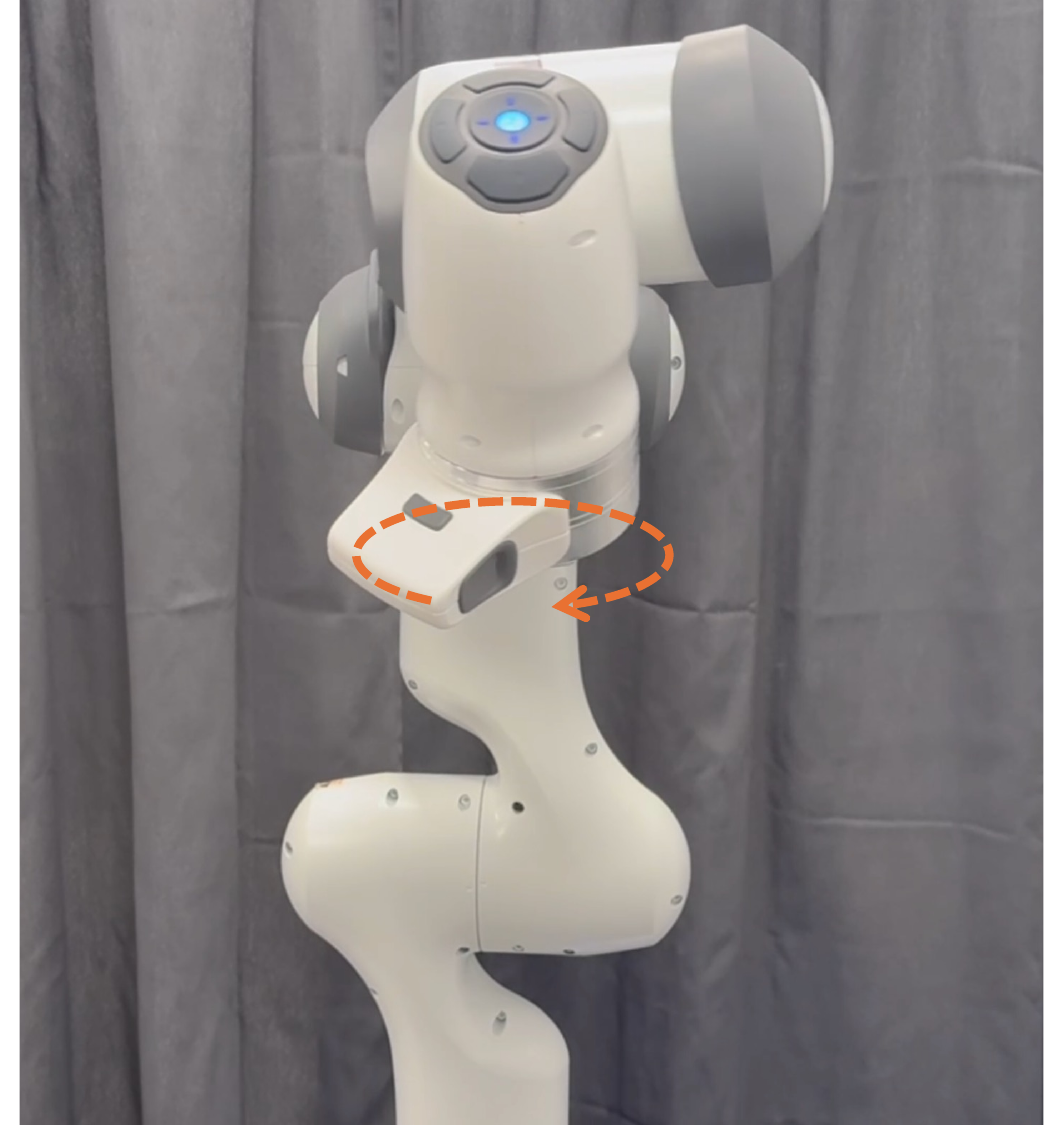}}\\
    \subfloat[]{
        \includegraphics[width=1\linewidth]{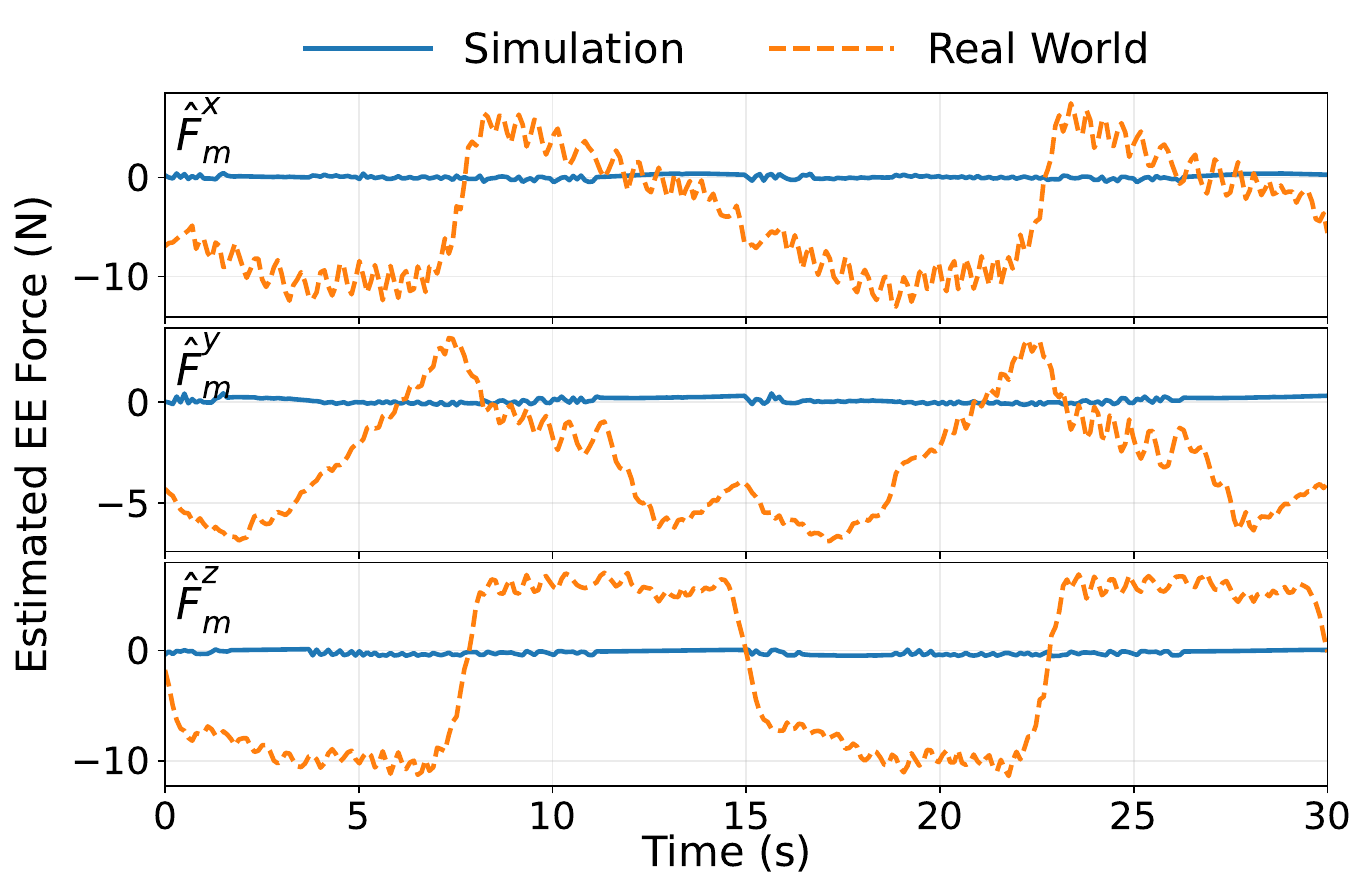}}
    \caption{Conventional momentum observer sim-to-real gap. (a) Test in Gazebo. (b) Test in real world. (c) Observer output difference between the simulation and real world.}
    \label{fig:Sim_to_Real} 
\end{figure}

The lumped disturbance $\hat{\bm{W}}_m \in \mathbb{R}^{6}$ at the end-effector can then be estimated as
\begin{equation}
    \hat{\bm{W}}_m = J(\bm{q})^{-T}\bm{r},
    \label{eq:F_ext_hat}
\end{equation}
where $\hat{\bm{W}}_{m}=\begin{bmatrix}\hat{\bm{F}}_{m}, \hat{\bm{\tau}}_{m}\end{bmatrix}^T$, $\hat{\bm{F}}_{m} \in \mathbb{R}^{3}$ denotes the estimated external force in translational space and $\hat{\bm{\tau}}_{m} \in \mathbb{R}^{3}$ represents estimated external torque in rotational space. However, in practical applications, the conventional momentum observer cannot reliably replace a force sensor due to the joint friction and unmodeled dynamics. As illustrated in Fig.~\ref{fig:Sim_to_Real}(a)(b), the manipulator in simulation and in the real world moves along the same circular trajectory without any external wrench so the desired $\hat{F}_m=0$. Fig.~\ref{fig:Sim_to_Real}(c) shows that the observer performs well in the simulation because of the ideal environmental conditions. However, the observer output in the real world is highly affected by the manipulator posture. To address this limitation, an NN is introduced to compensate for this sim-to-real gap. 

\subsection{Proposed Observer}
An effective approach to enhance the observer performance is to add an NN to compensate for the error. In this paper, force-related tasks are in translational space so the proposed observer is designed for estimating the translational force. Note that the proposed framework can be extended to rotational torque estimation if a torque sensor is available for data collection and validation. An NN $\bm{\Phi}(\bm{q}, \bm{\dot{q}})$ to compensate for force estimation is designed as
\begin{equation}
\bm{\hat{F}}_{\Phi(q, \dot{q})} = \bm{\hat{F}}_{m} + \bm{\Phi}(\bm{q}, \bm{\dot{q}}).
\label{eq:NN_F_ext_hat}
\end{equation}

Even though $\bm{\Phi}(\bm{q}, \bm{\dot{q}})$ can compensate for the modeling error, it can still be affected by the human applied disturbance. Human applied disturbance may change the joint velocity $\bm{\dot{q}}$ and then influence the output of $\bm{\Phi}(\bm{q}, \bm{\dot{q}})$. To address this, the joint velocity $\bm{\dot{q}}$ can be decomposed into task-space and null-space components. Firstly, the dynamically consistent generalized inverse of the Jacobian $\bar{\bm{J}}(\bm{q})$~\cite{Featherstone_1997_IJRR} is defined as
\begin{equation}
\bar{\bm{J}}(\bm{q})
=
\mathcal{M}^{-1}(\bm{q})\bm{J}^{T}(\bm{q})
\left[
\bm{J}(\bm{q})
\mathcal{M}^{-1}(\bm{q})
\bm{J}^{T}(\bm{q})
\right]^{-1}.
\label{eq:dynamic_consistent_inverse}
\end{equation}

According to \eqref{eq:dynamic_consistent_inverse}, the null-space projection matrix $\bm{N}$ can be defined as
\begin{equation}
\bm{N}
=
\bm{I}_{7}
-
\bar{\bm{J}}(\bm{q})\bm{J}(\bm{q}).
\label{eq:null_space_projection}
\end{equation}

Using this projection, the measured joint velocity can be decomposed as task-space component $\dot{\bm{q}}_t$ and null-space component $\dot{\bm{q}}_n$ where
\begin{align}
\dot{\bm{q}}_t
&=
\bar{\bm{J}}(\bm{q})\bm{J}(\bm{q})\dot{\bm{q}},\\
\dot{\bm{q}}_n
&=
\bm{N}\dot{\bm{q}}=
\left(
\bm{I}_{7}
-
\bar{\bm{J}}(\bm{q})\bm{J}(\bm{q})
\right)\dot{\bm{q}}.
\label{eq:task_velocity}
\end{align}

The task-space component $\dot{\bm{q}}_t$ represents the portion of the joint motion corresponding to the end-effector motion. In contrast, $\dot{\bm{q}}_n$ represents the internal motion within the kinematic redundancy of the manipulator. Since $\bm{J}(\bm{q})\dot{\bm{q}}_n = \bm{J}(\bm{q})\bm{N}\dot{\bm{q}} = \bm{0}$, the null-space component does not contribute to the end-effector velocity. Therefore, $\dot{\bm{q}}_t$ is applied instead of the complete joint velocity $\dot{\bm{q}}$ as the NN $\bm{\Phi}$ input, reducing the influence of disturbance-induced joint motion on the learned force estimation compensation. Eventually, \eqref{eq:NN_F_ext_hat} is replaced by the proposed force estimation defined as
\begin{equation}
\bm{\hat{F}}_{\Phi(q, \dot{q}_t)} = \bm{\hat{F}}_{m} + \bm{\Phi}(\bm{q}, \bm{\dot{q}}_t).
\label{eq:Final_F_ext_hat}
\end{equation}

In the training process, the input to the NN $\bm{\Phi}$ is $\bm{x} = [\bm{q},\bm{\dot{q}}_t]$, which is a 14-dimensional vector, while the network output is $\bm{y} = [\bm{F}_s - \bm{\hat{F}}_m]$, a 3-dimensional vector representing the force estimation error between sensor force $\bm{F}_s$ and the momentum observer force $\bm{\hat{F}}_m$. The neural network consists of three fully connected layers with an architecture of $14 \rightarrow 128 \rightarrow 128 \rightarrow 3$. The rectified linear unit (ReLU) is adopted as the activation function for all hidden layers. The learning framework is implemented in Python using PyTorch.

The proposed observer operates without requiring joint torque sensors and task-space force sensors during the real-time force estimation, thereby enabling a sensorless observer design. The force sensor mounted at the end-effector is used for offline data collection and result validation only and not used for force measurements in the experiment.

\section{Controller Design}
\subsection{Admittance Control}
To improve the compliance of the end-effector during contact interaction, an one-dimensional admittance controller is introduced along the end-effector normal direction. Let $\hat{F}^z_{\Phi(q, \dot{q}_t)}$ denote the z-axis component of the estimated force $\bm{\hat{F}}_{\Phi(q, \dot{q}_t)}$. The desired contact force is denoted by $F_d$. In this work, the admittance displacement is generated according to the second-order mass-spring-damper system:
\begin{equation}
    M_a \ddot{x}_a + D_a \dot{x}_a + K_a x_a = \hat{F}^z_{\Phi(q, \dot{q}_t)}  - F_d,\label{eq:mass-spring-damper}
\end{equation}
where $x_a$, $\dot{x}_a$, and $\ddot{x}_a$ are the admittance position, velocity, and acceleration, respectively. The parameters $M_a$, $D_a$, and $K_a$ represent the virtual admittance mass, damping, and stiffness. In the implementation, these parameters are selected as $M_a = 0.5, K_a = 100,$ and $ D_a = 14.14$.

According to \eqref{eq:mass-spring-damper}, the admittance acceleration $\ddot{x}_a$ is therefore computed as
\begin{equation}
    \ddot{x}_a
    =
    \frac{
    \hat{F}^z_{\Phi(q, \dot{q}_t)}  - F_d
    - D_a \dot{x}_a
    - K_a x_a
    }{M_a}.
\end{equation}

For discrete-time implementation, the admittance velocity $\dot{x}_a$ and position $x_a$ are updated using an integration operation as
\begin{align}
\dot{x}_a(t+1)
&=
\dot{x}_a(t)+\Delta t\,\ddot{x}_a(t),
\label{eq:admittance_velocity_update}
\\
x_a(t+1)
&=
x_a(t)+\Delta t\,\dot{x}_a(t+1),
\label{eq:admittance_position_update}
\end{align}
where $\Delta t$ denotes the control sampling interval. To ensure safe robot motion and prevent excessive compliant displacement, the maximum admittance position is limited to $0.02~\mathrm{m}$.

The resulting admittance displacement is applied along the desired Cartesian interaction direction. Let $\bm{R}\in SO(3)$ denote the rotation matrix of the end-effector with respect to the base frame, and let $\bm{n}=\bm{R}[0,0,1]^T$ denote the interaction direction along the end-effector z-axis expressed in the base frame. Since the force sensor is mounted with a fixed offset $l_s=0.09~\mathrm{m}$ along the end-effector normal direction, this geometric offset is incorporated into the end-effector reference position. The compliant Cartesian reference position is therefore generated as
\begin{equation}
\bm{p}_{\mathrm{ref}}
=
\bm{p}_d
-
(l_s+x_a)\bm{n},
\label{eq:admittance_reference}
\end{equation}
where $\bm{p}_d\in\mathbb{R}^3$ denotes the nominal desired position of the interaction point. In this way, the fixed sensor offset $l_s$ and the admittance displacement $x_a$ are both compensated along the interaction direction, while the end-effector remains the controlled Cartesian point. The admittance controller therefore modifies only the reference motion along the interaction direction without affecting the nominal trajectory in the orthogonal directions.

\subsection{Task-Space Pose and Force Controller}
To simultaneously achieve Cartesian pose tracking and interaction-force regulation, a task-space pose and force controller is developed. The position and orientation tracking controller generates a six-dimensional (6D) Cartesian control input, while an additional force control input is applied along the interaction direction $\bm{n}$. The two control inputs are combined in the task-space and mapped to the joint-space.

The Cartesian position error is defined as
\begin{equation}
\bm{e}_p
=
\bm{p}-\bm{p}_{\mathrm{ref}}.
\label{eq:position_error}
\end{equation}

For orientation tracking, let $\bm{\mathcal{Q}}$ and $\bm{\mathcal{Q}}_d$ denote the current and desired unit quaternions, respectively. The orientation error quaternion is defined as
\begin{equation}
\bm{\mathcal{Q}}_e
=
\bm{\mathcal{Q}}^{-1}
\otimes
\bm{\mathcal{Q}}_d,
\label{eq:orientation_error_quaternion}
\end{equation}
where $\otimes$ denotes quaternion multiplication. The orientation error used in the controller is obtained from the vector part of the error quaternion and expressed in the base frame as
\begin{equation}
\bm{e}_o
=
-\bm{R}\,
\mathrm{vec}\left(\bm{\mathcal{Q}}_e\right),
\label{eq:orientation_error}
\end{equation}
where $\mathrm{vec}(\cdot)$ extracts the vector part of a quaternion. The complete Cartesian pose tracking error is then defined as
\begin{equation}
\bm{e}
=
\begin{bmatrix}
\bm{e}_p\\
\bm{e}_o
\end{bmatrix}
\in\mathbb{R}^{6}.
\label{eq:cartesian_error}
\end{equation}

The desired Cartesian velocity is written as
\begin{equation}
\dot{\bm{x}}_d
=
\begin{bmatrix}
\dot{\bm{p}}_{\mathrm{ref}}\\
\bm{\omega}_d
\end{bmatrix},
\end{equation}
where $\bm{\omega}_d$ denotes the desired angular velocity. Accordingly, the Cartesian error derivative is
\begin{equation}
\dot{\bm{e}}
=
\bm{J}(\bm{q})\dot{\bm{q}}
-
\dot{\bm{x}}_d.
\label{eq:cartesian_error_dot}
\end{equation}

A nonsingular terminal sliding-type variable is constructed as
\begin{equation}
    \bm{s} = \bm{e} + \beta \bm{\dot{e}}^{\alpha},\label{eq:sliding_variable}
\end{equation}
where $\beta > 0$, $\alpha = p/q$, and $p > 0$ and $q > 0$ are adjacent odd numbers such that $1 < \alpha < 2$. In this work, $\alpha=5/3$ and $\beta=0.2$.

The Cartesian position and orientation control input is designed as
\begin{equation}
\begin{split}
\bm{u}_{\mathrm{pos}}
=&\,
\mathcal{\bar{M}}(\bm{q})
\left[
\ddot{\bm{x}}_d
-
\frac{
\dot{\bm{e}}^{2-\alpha}
}{
\alpha\beta
}
-
\bm{\Gamma}_1 \textrm{sgn}(\bm{s})
\right]
-
\bm{\Gamma}_2 \textrm{sgn}(\bm{s}),
\end{split}
\label{eq:position_orientation_controller}
\end{equation}
where $\bm{\Gamma}_1$ and $\bm{\Gamma}_2$ are positive control-gain matrices. The feedback control component of the proposed controller is inspired by the non-singular terminal sliding mode (NTSM) controller presented in~\cite{Feng_2002_Automatica}. The mass matrix in Cartesian space $\mathcal{\bar{M}}(\bm{q})$ significantly reduces the control effect on the last three joints of the manipulator. Therefore, an additional feedback term $-\bm{\Gamma}_2 \textrm{sgn}(\bm{s})$ is introduced to compensate for this effect.

A force regulation is performed along the end-effector interaction direction $\bm{n}$. The force tracking error is defined as
\begin{equation}
e_F
=
\hat{F}^z_{\Phi(q, \dot{q}_t)} - F_d.
\label{eq:force_tracking_error}
\end{equation}

A proportional-derivative (PD) controller is employed to generate the commanded interaction force defined as
\begin{equation}
F_\mathrm{cmd}
=
F_d
+
K_{f}e_F
+
D_{f}\dot{e}_F,
\label{eq:force_controller}
\end{equation}
where $K_{f}$ and $D_{f}$ denote the proportional and derivative force control gains, respectively. In this work, $K_{f}=-3$ and $D_{f}=-0.003$.

Since a force control is applied only along the interaction direction $\bm{n}$, its corresponding 6D Cartesian input is expressed as
\begin{equation}
\bm{u}_{\mathrm{force}}
=
\begin{bmatrix}
F_\mathrm{cmd}\bm{n}\\
\bm{0}_{3\times1}
\end{bmatrix}.
\label{eq:force_control_input}
\end{equation}

Finally, the pose tracking and force control joint torque commands are combined as
\begin{equation}
\bm{\tau}_\mathrm{task} = J(\bm{q})^{T}(\bm{u}_{\mathrm{pos}} + \bm{u}_{\mathrm{force}}) + \mathcal{C}(\bm{q},\bm{\dot{q}})\bm{\dot{q}} + \mathcal{G}(\bm{q}).
\label{eq:task_torque}
\end{equation}

Therefore, the proposed task-space controller simultaneously regulates the Cartesian position and orientation while controlling the interaction force along the contact direction.

\subsection{Null-Space Compliant Control}

For a redundant manipulator, the task-space control objective does not uniquely determine the joint motion. Therefore, the remaining redundancy can be utilized to calculate the robot posture without affecting the end-effector task. In this work, a null-space compliant controller is introduced to improve the compliance of the robot.

A virtual spring-damper controller is designed in the null-space. Let $\bm{q}_{d,n} \in \mathbb{R}^{7}$ denote the desired null-space joint posture. The raw null-space torque is defined as
\begin{equation}
    \bm{\tau}_\mathrm{null,raw}
    =
    \bm{K}_n
    (\bm{q}_{d,n} - \bm{q})
    -
    \bm{D}_n
    \bm{\dot{q}}_n,\nonumber
\end{equation}
where $\bm{K}_n$ and $\bm{D}_n$ are the null-space stiffness and damping coefficients, respectively. The damping coefficient is selected according to the relationship $\bm{D}_n = 2\sqrt{\bm{K}_n}$.

Then the null-space torque applied to the robot is obtained by projecting the raw torque into the null-space as
\begin{equation}
    \bm{\tau}_\mathrm{null}
    =
    \bm{N}^{T}
    \bm{\tau}_\mathrm{null,raw}.\nonumber
\end{equation}

Finally, the commanded control torque can be written as
\begin{equation}
\begin{split}
    \bm{\tau}_\mathrm{cmd} &= \bm{\tau}_\mathrm{task} + \bm{\tau}_\mathrm{null}\\
    &= \underbrace{J(\bm{q})^{T}\bm{u}_{\mathrm{pos}}}_{\text{task-space pose}} + \underbrace{J(\bm{q})^{T}\bm{u}_{\mathrm{force}}}_{\text{task-space force}}\\\
    &+ \underbrace{\bm{N}^{T}\bm{\tau}_\mathrm{null,raw}}_{\text{null-space}}+\underbrace{\mathcal{C}(\bm{q},\bm{\dot{q}})\bm{\dot{q}} + \mathcal{G}(\bm{q})}_{\text{feedforward}}.
\end{split}\nonumber
\end{equation}

\begin{figure*}[t]
    \centering
    \includegraphics[width=1\linewidth]{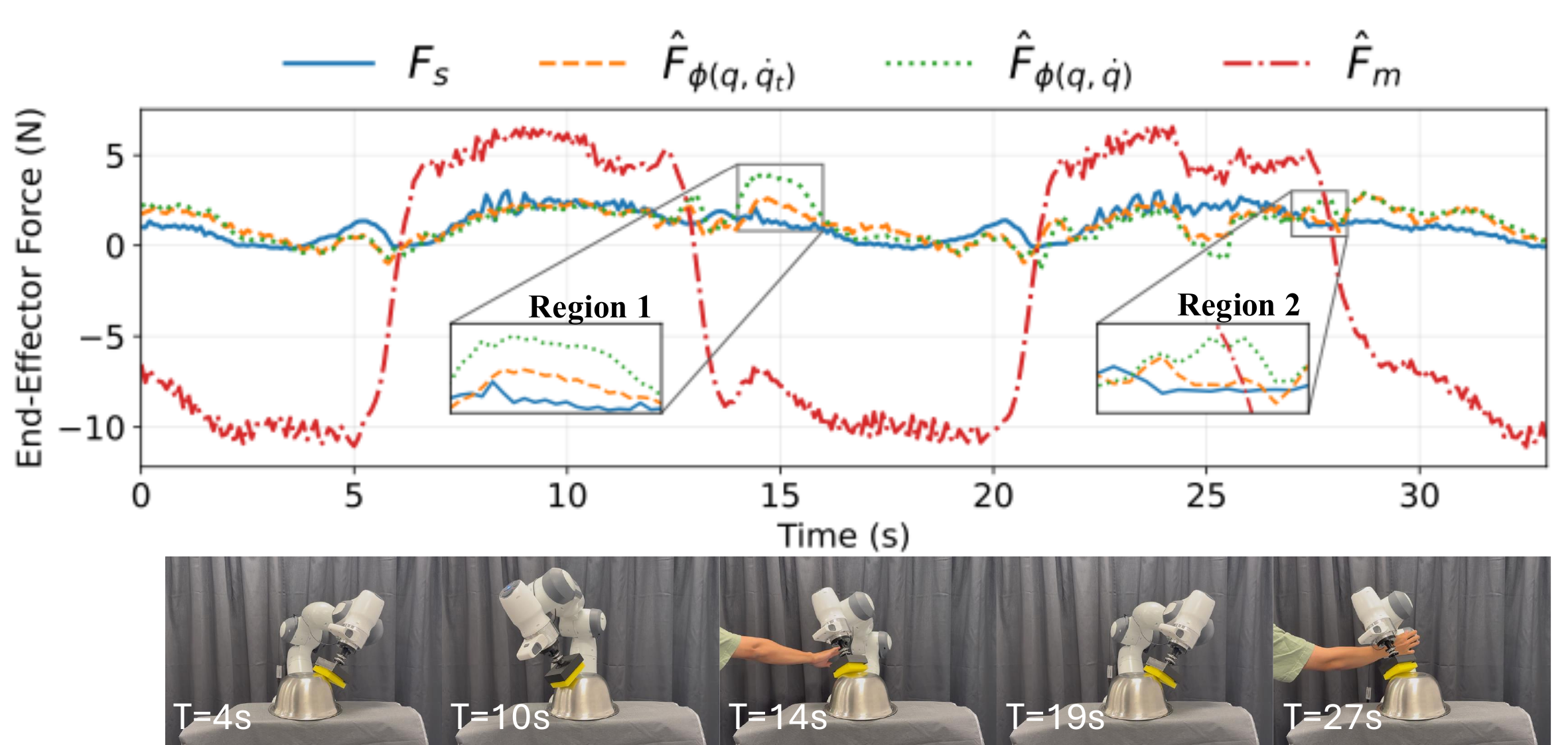}
    \caption{Comparison of the end-effector force estimation performance under human applied disturbances. The conventional momentum observer, the learning-based observer using the full joint velocity, and the proposed observer were compared with sensor force. The enlarged regions within box lines highlight the estimation performance during human disturbance periods at around $T=14 \ s$ and $27 \ s$ respectively.}
    \label{fig:Disturbance}
\end{figure*}
\subsection{Stability Analysis}
With the nominal Coriolis and gravity terms compensated by the feedforward terms, the closed-loop task-space dynamics \eqref{dynamic_cs} can be written as
\begin{equation}
\mathcal{\bar{M}}(\bm{q})\bm{\ddot{x}}
=
\bm{u}_{\mathrm{pos}}
+
\bm{u}_{\mathrm{force}}
+
\bm{u}_d.
\label{eq:closed_loop_task_dynamics}
\end{equation}

Define $\bm{D}_{e} = \operatorname{diag}( |\dot e_1|^{\alpha-1}, \ldots, |\dot e_6|^{\alpha-1})$. The derivative of the sliding surface \eqref{eq:sliding_variable} is then given by
\begin{equation}
\bm{\dot{s}}
=
\bm{\dot{e}}
+
\alpha\beta
\bm{D}_{e}
\bm{\ddot e}.
\label{eq:s_dot}
\end{equation}

Substituting the pose controller \eqref{eq:position_orientation_controller} into \eqref{eq:closed_loop_task_dynamics}, and using
$\bm{\ddot e}=\bm{\ddot x}-\bm{\ddot x}_d$, gives
\begin{equation}
\begin{split}
\bm{\ddot e}
=&
-
\frac{\bm{\dot e}^{2-\alpha}}{\alpha\beta}
-
\bm{\Gamma}_1 \operatorname{sgn}(\bm{s})
-
\mathcal{\bar M}^{-1}\bm{\Gamma}_2
\operatorname{sgn}(\bm{s})
\\
&+
\mathcal{\bar M}^{-1}
\left(
\bm{u}_{\mathrm{force}}+\bm{u}_d
\right).
\end{split}
\label{eq:error_dynamics}
\end{equation}

Substituting \eqref{eq:error_dynamics} into \eqref{eq:s_dot} gives
\begin{equation}
\begin{split}
\bm{\dot{s}}
=&
\bm{\dot e}
-
\bm{D}_{e}\bm{\dot e}^{2-\alpha}
-
\alpha\beta
\bm{D}_{e}
\bm{\Gamma}_1
\operatorname{sgn}(\bm{s})
\\
&-
\alpha\beta
\bm{D}_{e}
\mathcal{\bar M}^{-1}
\bm{\Gamma}_2
\operatorname{sgn}(\bm{s})
\\
&+
\alpha\beta
\bm{D}_{e}
\mathcal{\bar M}^{-1}
\left(
\bm{u}_{\mathrm{force}}+\bm{u}_d
\right).
\end{split}
\label{eq:s_dot_1}
\end{equation}

Since $\bm{D}_{e}\bm{\dot e}^{2-\alpha} = \bm{\dot e}$, the first two terms cancel. Therefore, \eqref{eq:s_dot_1} can be rewritten as
\begin{equation}
\begin{split}
\bm{\dot{s}}
=&
-
\alpha\beta
\bm{D}_{e}
\left(
\bm{\Gamma}_1
+
\mathcal{\bar M}^{-1}\bm{\Gamma}_2
\right)
\operatorname{sgn}(\bm{s})
\\
&+
\alpha\beta
\bm{D}_{e}
\mathcal{\bar M}^{-1}
\left(
\bm{u}_{\mathrm{force}}+\bm{u}_d
\right).
\end{split}
\label{eq:sliding_dynamics}
\end{equation}

Consider the Lyapunov candidate \cite{Shevitz_1994_TAC}
\begin{equation}
V
=
\sum_{i=1}^{6}|s_i|,
\end{equation}
which is positive definite with respect to $\bm{s}$ and its time derivative is
\begin{equation}
\dot V
=
\operatorname{sgn}(\bm{s})^{T}\bm{\dot{s}}.
\label{eq:V_dot}
\end{equation}

Let $\bm{\Delta}_u=\bm{u}_{\mathrm{force}}+\bm{u}_d$ denote the residual input mismatch after disturbance compensation, where $\bm{u}_{\mathrm{force}}$ is the force-related input computed based on the desired interaction force and the estimated interaction force, and $\bm{u}_d$ represents the unknown disturbance. Under ideal compensation, the disturbance-related component of $\bm{u}_{\mathrm{force}}$ exactly compensates for $\bm{u}_d$. In practice, exact cancellation cannot be guaranteed due to force-estimation errors, modeling uncertainties, and other unmodeled effects.Therefore, $\bm{\Delta}_u$ captures the remaining compensation error, and the corresponding residual compensation disturbance is defined as $\bm{d}=\bar{\mathcal{M}}^{-1}\bm{\Delta}_u$. Assume that the residual compensation disturbance is uniformly bounded such that
\begin{equation}
\left\|
\bm{d}
\right\|_{\infty}
=
\left\|
\mathcal{\bar M}^{-1}\bm{\Delta}_u
\right\|_{\infty}
\leq
\bar d,
\label{eq:disturbance_bound}
\end{equation}
where $\bar d>0$ is a known constant.

Substituting \eqref{eq:sliding_dynamics} into
\eqref{eq:V_dot} gives
\begin{equation}
\begin{split}
\dot V
=&
-
\alpha\beta
\operatorname{sgn}(\bm{s})^{T}
\bm{D}_{e}
\left(
\bm{\Gamma}_1
+
\mathcal{\bar M}^{-1}\bm{\Gamma}_2
\right)
\operatorname{sgn}(\bm{s})
\\
&+
\alpha\beta
\operatorname{sgn}(\bm{s})^{T}
\bm{D}_{e}
\bm{d}.
\end{split}
\label{eq:V_dot_disturbance}
\end{equation}

Using the disturbance bound in \eqref{eq:disturbance_bound},
the second term in \eqref{eq:V_dot_disturbance} satisfies
\begin{equation}
\begin{split}
\operatorname{sgn}(\bm{s})^{T}
\bm{D}_{e}\bm{d}
&\leq
\left|
\operatorname{sgn}(\bm{s})^{T}
\bm{D}_{e}\bm{d}
\right|
\\
&\leq
\left\|
\bm{D}_{e}\operatorname{sgn}(\bm{s})
\right\|_{1}
\left\|\bm{d}\right\|_{\infty}
\\
&\leq
\bar d
\left\|
\bm{D}_{e}\operatorname{sgn}(\bm{s})
\right\|_{1}.
\end{split}
\label{eq:disturbance_term_bound}
\end{equation}

Suppose that the control gains $\bm{\Gamma}_1$ and
$\bm{\Gamma}_2$ are selected such that there exists a constant
$\eta>0$ satisfying
\begin{equation}
\begin{split}
&
\operatorname{sgn}(\bm{s})^{T}
\bm{D}_{e}
\left(
\bm{\Gamma}_1
+
\mathcal{\bar M}^{-1}\bm{\Gamma}_2
\right)
\operatorname{sgn}(\bm{s})
\\
&\qquad\geq
(\bar d+\eta)
\left\|
\bm{D}_{e}\operatorname{sgn}(\bm{s})
\right\|_{1}.
\end{split}
\label{eq:gain_condition}
\end{equation}

Using \eqref{eq:disturbance_term_bound} and
\eqref{eq:gain_condition}, \eqref{eq:V_dot_disturbance} becomes 
\begin{equation}
\dot V
\leq
-
\alpha\beta\eta
\left\|
\bm{D}_{e}
\operatorname{sgn}(\bm{s})
\right\|_{1}
\leq
0,
\label{eq:V_dot_negative}
\end{equation}
for any nonzero velocity tracking errors. 

Therefore, the Lyapunov function is non-increasing under the bounded lumped disturbance, and the stabilizing control term dominates the effects of the force estimation errors, interaction disturbances, and unmodeled dynamics. As a result, the closed-loop sliding dynamics remain stable and satisfy the desired robustness condition. This concludes the stability analysis. \qed

\section{Performance Evaluation}
All experiments in this section are performed using a 7-DOF Franka Emika robotic manipulator with the custom-designed end-effector as shown in Fig.~\ref{fig:Hardware}(b). The experiments are conducted on a desktop equipped with an Intel i7 CPU and Intel UHD Graphics 630 GPU. All components in the proposed framework are executed in real time and the force sensor is not used for real-time feedback control. The observer network $\bm{\Phi}(\bm{q}, \bm{\dot{q}}_t)$ is integrated directly into the control loop. The total computation time is below 1\,ms, ensuring real-time performance without introducing additional control latency. The experimental videos are available at
\href{https://drive.google.com/drive/folders/1Gkz69RbWp93sgu5dC459l0ypgagU7j6w?usp=drive_link}{this link}.

\subsection{Estimation Results}
To evaluate the robustness of the proposed disturbance observer against human disturbances, experiments were conducted while the robot wiped the bowl surface and manual external disturbances from the human arm were applied to the robot body. During the wiping task, the robot tracks a desired force and a 6D pose and the disturbances were applied to the robot body. Since the interaction force is primarily applied along the end-effector z-axis in the wiping task, only the z-direction force is presented for comparisons. The end-effector force measured by the force sensor, denoted by $F_s^z$, was used as the reference for the comparison purpose. The conventional momentum observer $\hat{F}_{m}^z$, the learning-based observer using the full joint velocity $\hat{F}_{\Phi(\bm{q},\dot{\bm{q}})}^z$, and the proposed observer using the task-space joint velocity $\hat{F}_{\Phi(\bm{q},\dot{\bm{q}}_{t})}^z$ were compared.

In the training and validation processes, the dataset comprised 355K samples collected under applied forces of 0, 1, 2, 3, and 4~N with random human disturbance. For each force condition, the first 80\% of samples were used for training and the remaining 20\% for validation. Both $\Phi(\bm{q},\dot{\bm{q}})$ and $\Phi(\bm{q},\dot{\bm{q}}_{t})$ were trained for 50 epochs using the Adam optimizer with a learning rate of $10^{-3}$, a batch size of 512, and mean squared error (MSE) as the loss function. To ensure a fair comparison, the random seed was fixed at 42, and identical initial weights, data partitions, and minibatch ordering were used for both NNs. The NN $\Phi(\bm{q},\dot{\bm{q}}_{t})$ achieved validation MSE, root mean square error (RMSE), and mean absolute error (MAE) values of 1.2032, 1.0969, and 0.7931, respectively, representing reductions of 2.15\%, 1.08\%, and 7.21\% relative to the NN $\Phi(\bm{q},\dot{\bm{q}})$. Under the 1, 2, and 3~N conditions, RMSE decreased by 10.78\%, 12.43\%, and 16.92\%, respectively.

{\renewcommand{\arraystretch}{1.3}
\begin{center}
    \begin{table}[t]
        \caption{Comparisons of the end-effector force estimation performance}
        \label{tab:observer_performance}
        \resizebox{\columnwidth}{!}{%
        \begin{tabular}{ccccc}
            \toprule
            \multirow{2}{*}{\textbf{Observer}}
            & \textbf{Overall}
            & \textbf{Region 1}
            & \textbf{Region 2}
            & \textbf{Maximum}\\
            & \textbf{RMSE(N)}
            & \textbf{RMSE(N)}
            & \textbf{RMSE(N)}
            & \textbf{Error(N)}\\
            \midrule
            $\hat{F}_{m}^z$
            & 7.513
            & 9.334
            & 3.103
            & 12.253\\
            
            $\hat{F}_{\Phi(\bm{q},\dot{\bm{q}})}^z$
            & 0.960
            & 2.016
            & 0.878
            & 3.054\\
            
            $\bm{\hat{F}_{\Phi(\bm{q},\dot{\bm{q}}_{t})}^z}$
            & \textbf{0.729}
            & \textbf{0.804}
            & \textbf{0.408}
            & \textbf{1.892}\\
            \bottomrule
        \end{tabular}%
        }
    \end{table}
\end{center}
}

To further evaluate the proposed observer performance, a real time comparison is executed as shown in Fig.~\ref{fig:Disturbance}. The conventional momentum observer output $\hat{F}_{m}^z$ is strongly affected by the robot posture and disturbances applied to the robot body. Large deviations from the measured end-effector force can be observed. This is because the conventional observer estimates the overall joint-space disturbance and cannot effectively distinguish the joint friction and model errors from the end-effector force interaction.
\begin{figure*}[t]
    \centering
    \subfloat[]{
        \includegraphics[width=0.5\linewidth]{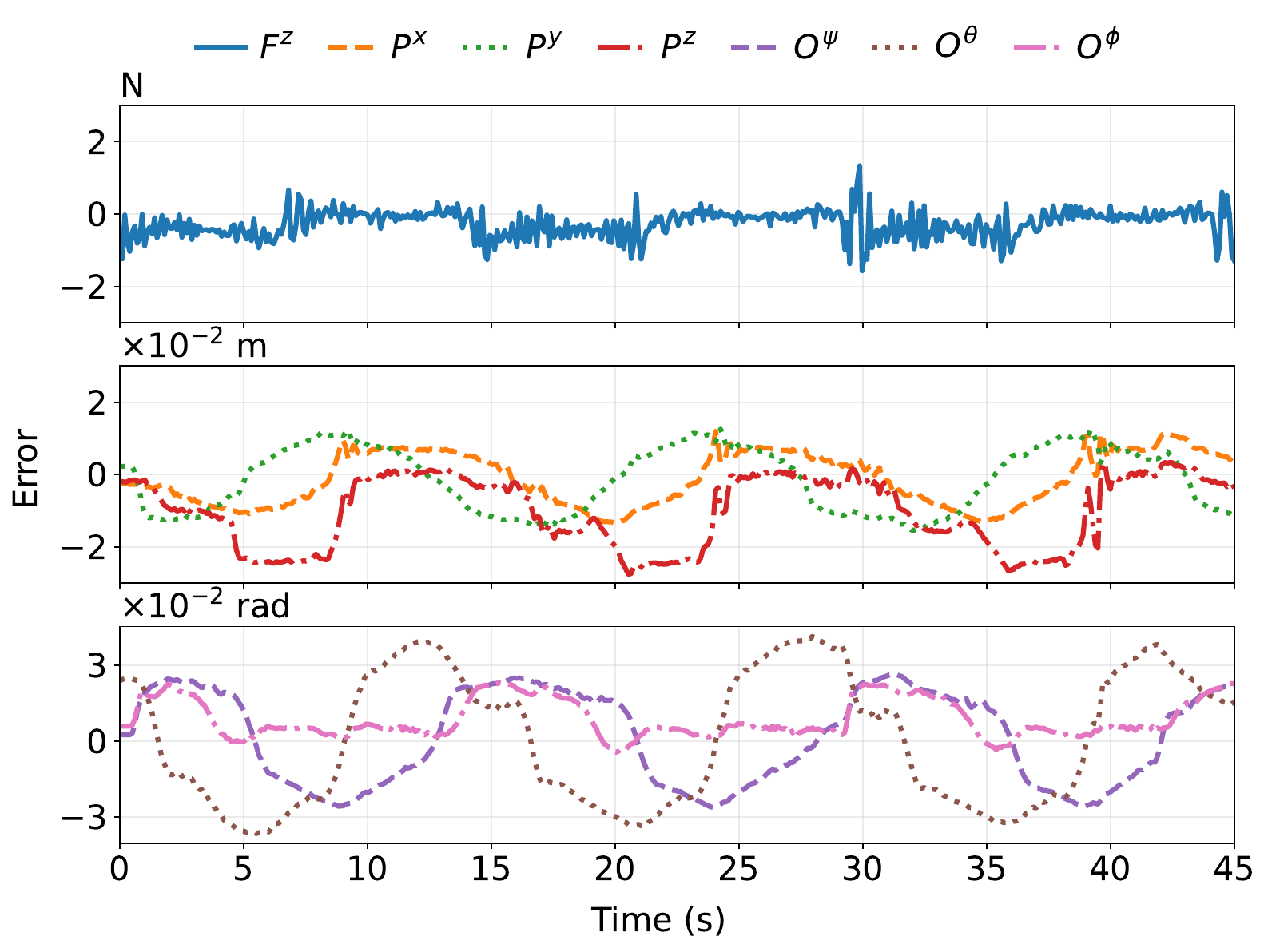}}
    \subfloat[]{
        \includegraphics[width=0.5\linewidth]{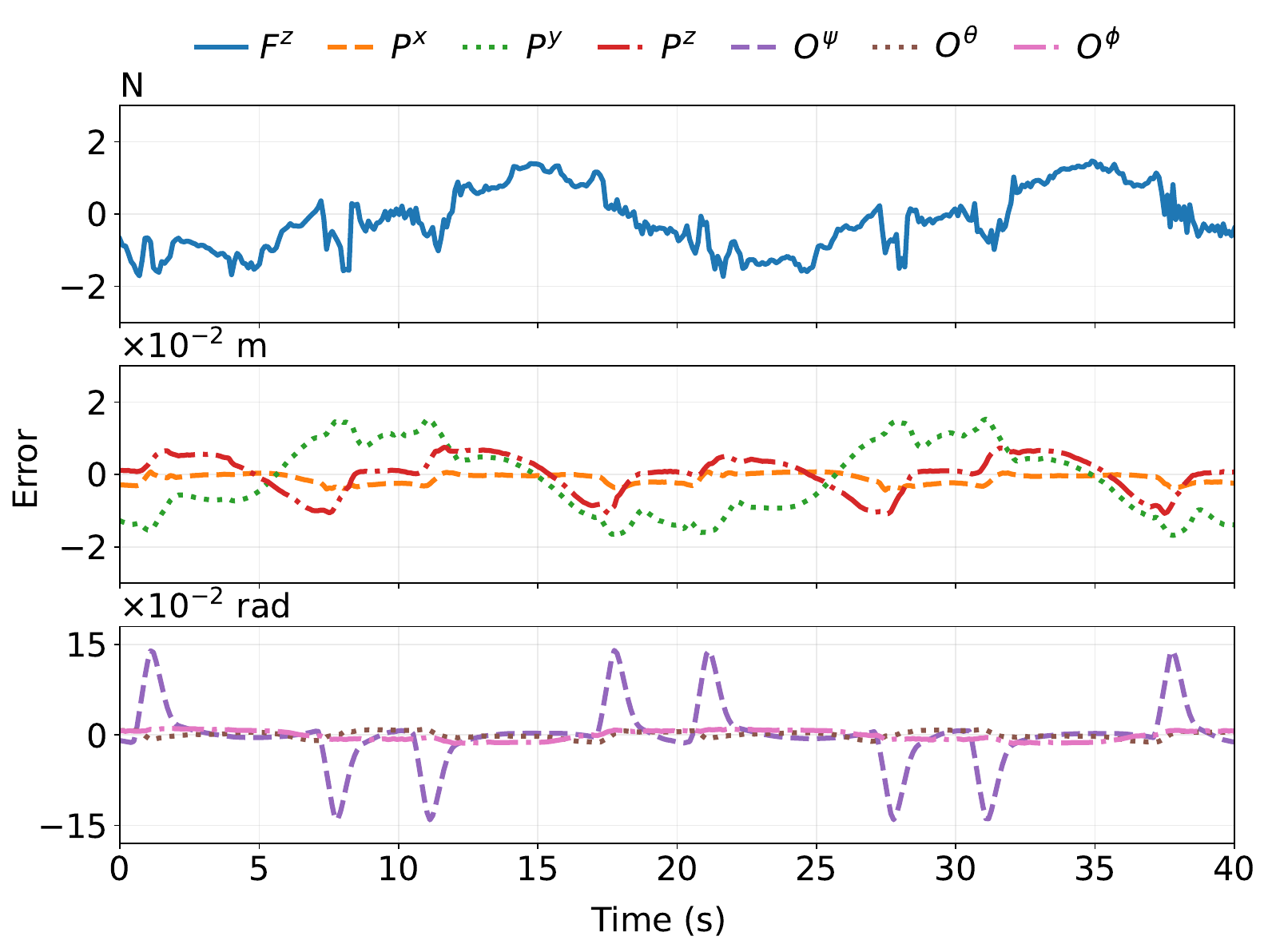}}
    \caption{Controller tracking performance with two reference trajectories (a) $R_1$ and (b) $R_2$. The translational position tracking errors $P^x$, $P^y$, and $P^z$ are along the x, y and z directions. The orientation tracking errors $O^{\psi}$, $O^{\theta}$, and $O^{\phi}$ are yaw, pitch, and roll. The force tracking errors $F^z$ are along the z direction.}
    \label{fig:Tracking} 
\end{figure*}

In contrast, both learning-based observers significantly reduce the influence of joint friction and model errors. They both provide force estimations that are significantly closer to the sensor measurement. As further illustrated by the enlarged regions at around $T=14 \ s$ and $27 \ s$ in  Fig.~\ref{fig:Disturbance}, $\hat{F}_{\Phi(\bm{q},\dot{\bm{q}})}^z$ follows the overall variation of $F_s^z$, but estimation deviations remain small during the motion under disturbances. That is because the NN uses the overall joint-space velocity so it cannot separate disturbance from the end-effector force interaction. By replacing the full joint velocity with its task-space component $\dot{\bm{q}}_{t}$, the proposed $\hat{F}_{\Phi(\bm{q},\dot{\bm{q}}_{t})}^z$ provides an estimate closer to $F_s^z$ under human disturbances.

For quantitative comparison, the maximum error and the RMSE of the estimated end-effector force are provided. To specifically evaluate the robustness of each observer against human disturbances, additional RMSEs are calculated using the samples within two human disturbance periods. The quantitative results are summarized in Table~\ref{tab:observer_performance}. Compared with the conventional momentum observer, both learning-based approaches substantially reduce the estimation error. In particular, the proposed $\hat{F}_{\Phi(\bm{q},\dot{\bm{q}}_{t})}^z$ achieves an overall RMSE of 0.729 N and a maximum error of 1.892 N. Compared with the learning-based observer $\hat{F}_{\Phi(\bm{q},\dot{\bm{q}})}^z$, the proposed method reduces the overall RMSE by 24.1\% and the maximum error by 38.0\%. More importantly, during two human disturbance periods, the RMSE is reduced from 2.016 N to 0.804 N and from 0.878 N to 0.408 N, corresponding to improvements of 60.1\% and 53.5\%, respectively. These results demonstrate that replacing the full joint velocity with its task-space component effectively reduces the influence of human disturbances on the estimated end-effector interaction force.

\subsection{Tracking Performance}
\begin{figure}[t]
    \centering
    \subfloat[]{
        \includegraphics[width=0.47\linewidth]{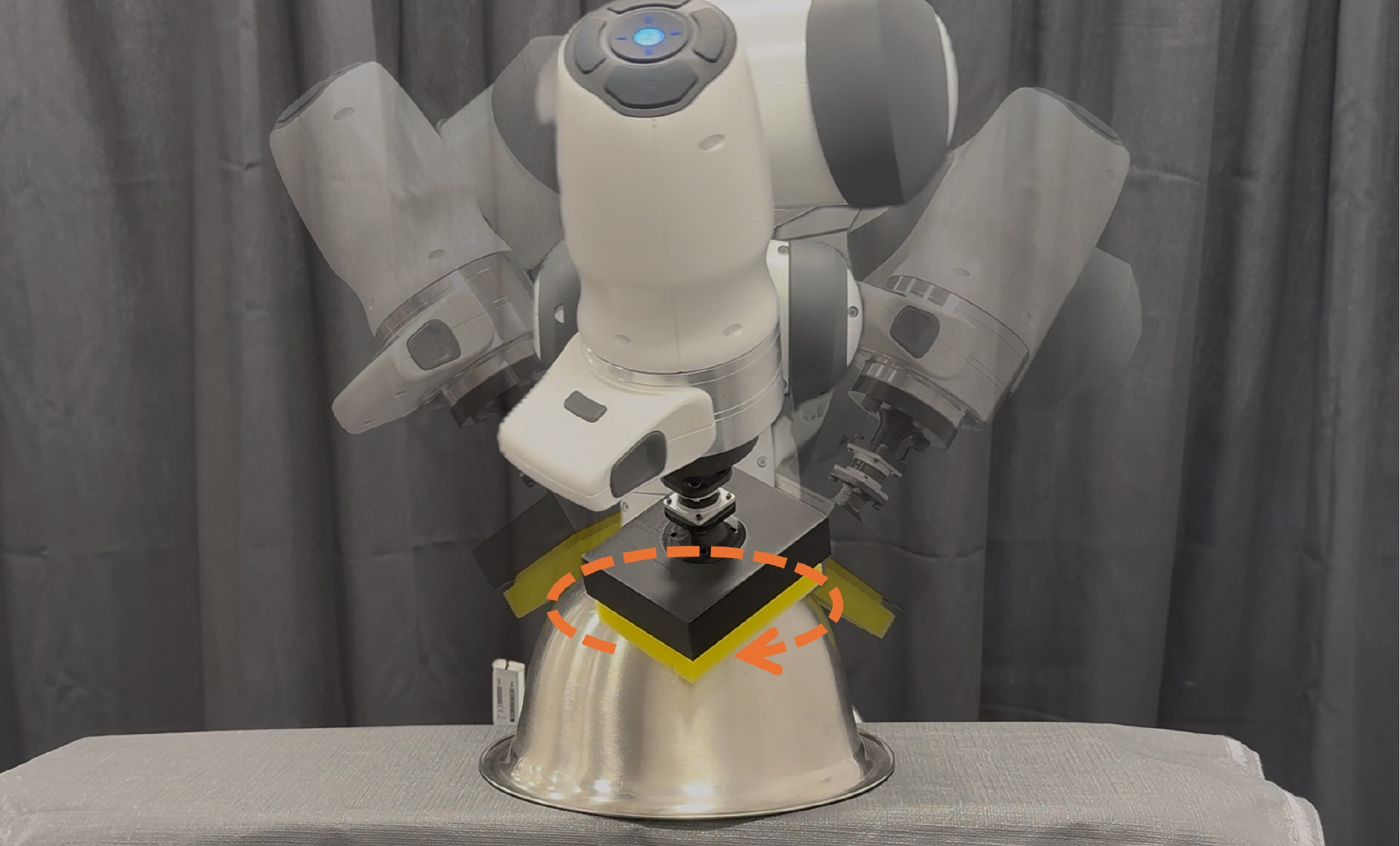}}
    \subfloat[]{
        \includegraphics[width=0.47\linewidth]{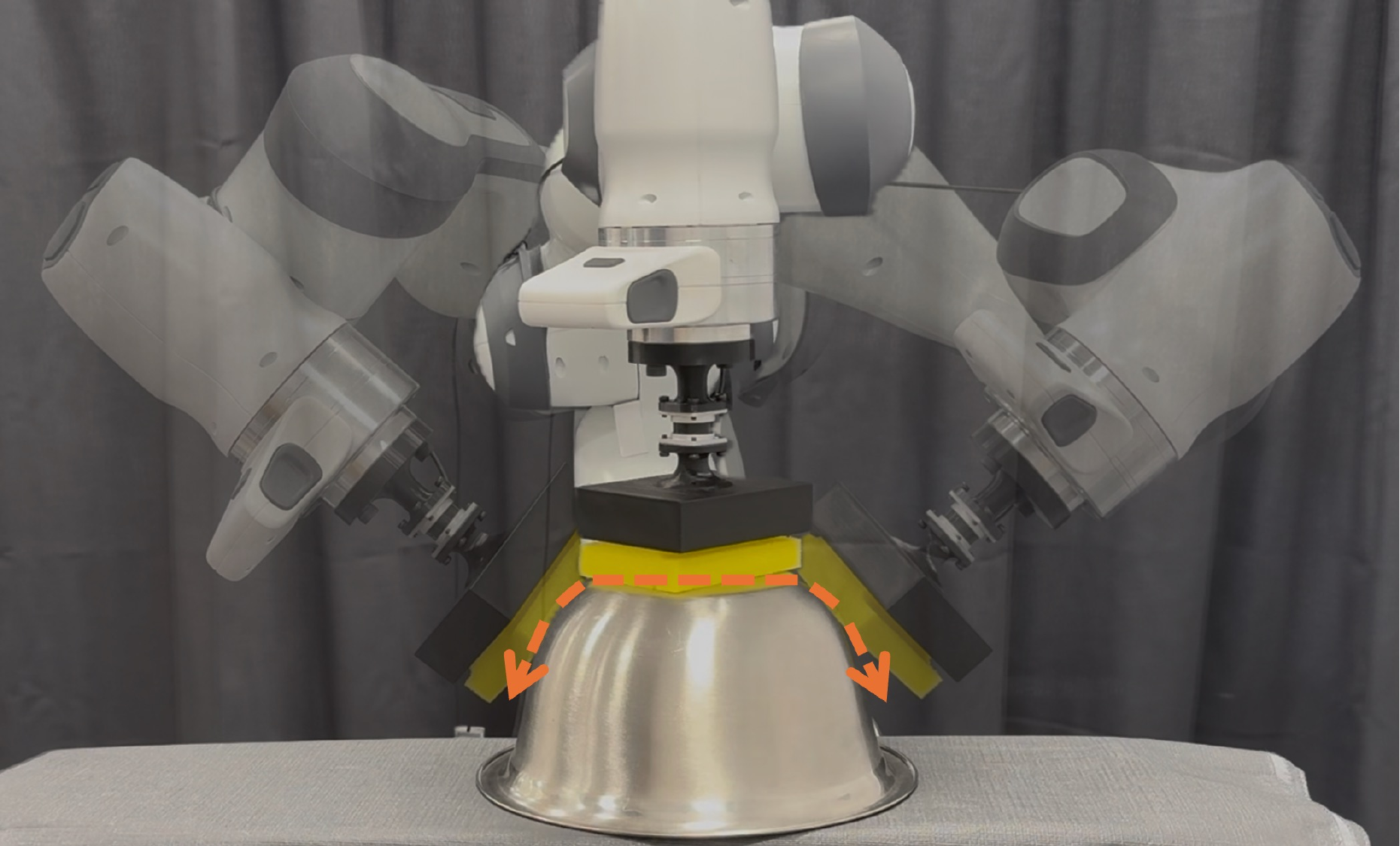}}
    \caption{Reference trajectories $R_1$ and $R_2$. (a) $R_1$ is a circular trajectory and (b) $R_2$ is a rounded trapezoidal trajectory.}
    \label{fig:Trajectory} 
\end{figure}
The tracking performance of the proposed controller is evaluated using two reference trajectories $R_1$ and $R_2$. As shown in Fig.~\ref{fig:Trajectory}(a), trajectory $R_1$ employs a circular trajectory in the x-y plane with a constant z position, while the end-effector orientation varies continuously during the motion. Trajectory $R_2$ employs a rounded trapezoidal trajectory in the y-z plane with a constant x position, while the end-effector orientation is adjusted along the trajectory as illustrated in Fig.~\ref{fig:Trajectory}(b). During both tasks, the controller simultaneously regulates the end-effector pose and the interaction force along the z direction. 

The corresponding pose and force tracking results are shown in Fig.~\ref{fig:Tracking}, while the quantitative tracking errors are summarized in Table~\ref{tab:tracking_rmse}. For trajectory $R_1$, the experimental results show that the desired position and orientation trajectories are closely tracked despite the simultaneous translational and rotational motion. The RMSE of the force tracking errors $F^z$ is 0.464 N. The position tracking RMSEs $P^x$, $P^y$, and $P^z$ are 0.732, 0.888, and 1.403 cm, respectively. Meanwhile, the orientation tracking RMSEs $O^{\psi}$, $O^{\theta}$, and $O^{\phi}$ are 0.018, 0.026, and 0.012 rad, respectively. For trajectory $R_2$, the controller is further evaluated with a different motion in the local trajectory direction. The proposed controller maintains stable force and pose tracking throughout the motion. The force tracking RMSE is 0.876 N, while the position tracking RMSEs are 0.179, 0.990, and 0.506 cm, respectively. The corresponding orientation tracking RMSEs are 0.047, 0.006, and 0.008 rad, respectively.

Overall, the position tracking RMSE remains below $1.5$~cm and the orientation tracking RMSE remains below $0.05$~rad for both tasks, while the force tracking RMSE remains below $0.9$~N. Although the two tasks exhibit different error distributions due to their distinct trajectory types and motion directions, consistently bounded tracking errors are achieved in both cases. These results demonstrate that the proposed controller can achieve accurate task-space pose tracking and interaction force regulation at the same time.
{\renewcommand{\arraystretch}{1.3}
\begin{table}[h]
    \centering
    \caption{Tracking Performance}
    \label{tab:tracking_rmse}
    \begin{tabular}{ccc}
        \toprule
        \textbf{Tracking Error}
        & \textbf{$R_1$ RMSE}
        & \textbf{$R_2$ RMSE} \\
        \midrule
        $F^{z}$ (N)
        & 0.464
        & 0.876 \\

        $P^{x}$ (cm)
        & 0.732
        & 0.179 \\

        $P^{y}$ (cm)
        & 0.888
        & 0.990 \\

        $P^{z}$ (cm)
        & 1.403
        & 0.506 \\

        $O^{\psi}$ (rad)
        & 0.018
        & 0.047 \\

        $O^{\theta}$ (rad)
        & 0.026
        & 0.006 \\

        $O^{\phi}$ (rad)
        & 0.012
        & 0.008 \\
        \bottomrule
    \end{tabular}
\end{table}
}

\section{Conclusion}
This work developed an observer and controller framework for both task-space robust tracking and null-space compliant interaction. By integrating an NN and decomposing the joint velocity into task-space and null-space components, the proposed observer is less affected by human disturbances applied to the robot body. The proposed observer has an overall RMSE of 0.729 N. Moreover, in the context of human disturbance, the proposed observer reduces the RMSE by 60.1\% and 53.5\% compared with the learning-based observer. Based on the estimated interaction force, the proposed controller further achieved simultaneous task-space pose tracking, force regulation, and compliant null-space behavior.

The main limitations of the current work are the dependence of the learned observer on offline training data and the relatively limited range of experimental trajectories and disturbance conditions. Future work will investigate adaptive and online learning to improve generalization under varying interaction conditions, and evaluate the proposed method in more complex and unstructured pHRI tasks.


\bibliographystyle{Bibliography/IEEEtranTIE}
\bibliography{Bibliography/IEEEabrv,Bibliography/BIB_25-TMech-xxxx}\ 
\newpage
\vspace{-1cm}
\begin{IEEEbiography}
[{\includegraphics[width=1in,height=1.25in,clip,keepaspectratio]{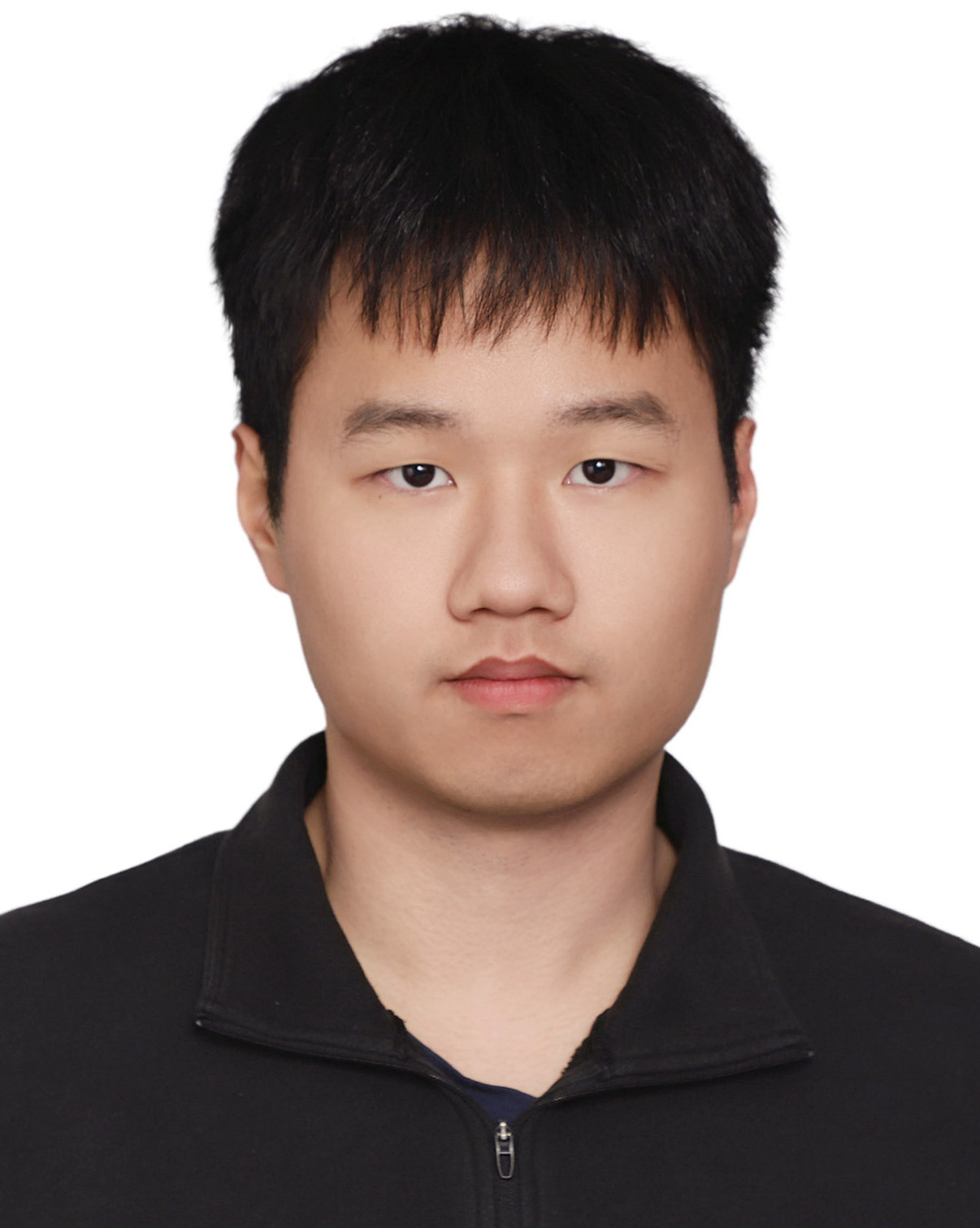}}]{Nuo Chen} (Student Member, IEEE) received the B.E. degree in robotics engineering from Southern University of Science and Technology, Shenzhen, China, in 2022, and M.A.Sc. degree in mechanical engineering from Dalhousie University, Halifax, NS, Canada, in 2024. He is currently working toward the Ph.D. degree in mechanical engineering with Dalhousie University, Halifax, NS, Canada.

His research interests include robotics control and machine learning.
\end{IEEEbiography}
\vspace{-1cm}
\begin{IEEEbiography}[{\includegraphics[width=1in,height=1.25in,clip,keepaspectratio]{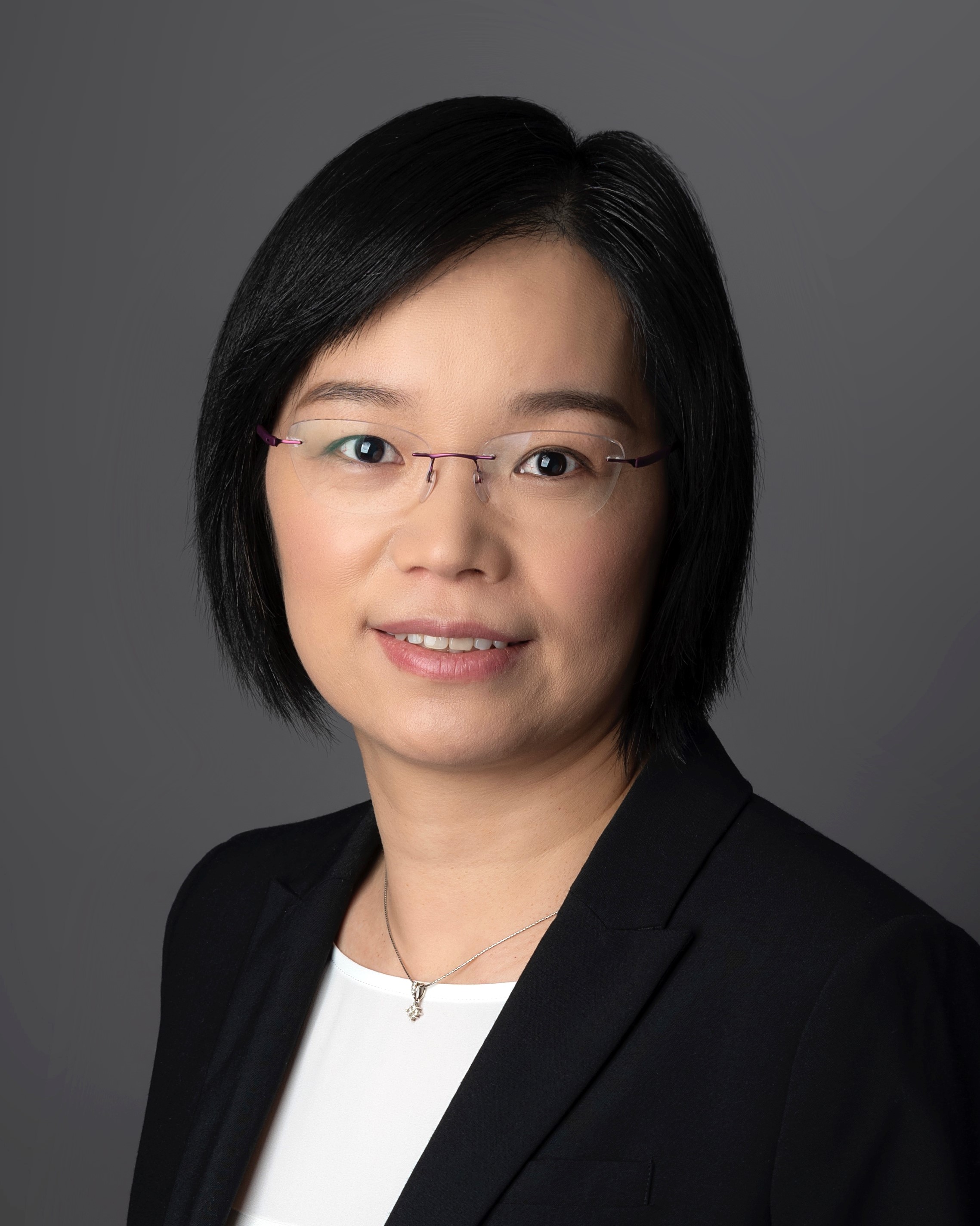}}]{Ya-Jun Pan} (Senior Member, IEEE) received the B.E. degree in mechanical engineering from Yanshan University, Qinhuangdao, China, in 1996, the M.E. degree in mechanical engineering from Zhejiang University, Hangzhou, China, in 1999, and the Ph.D. degree in electrical and computer engineering from the National University of Singapore, Singapore, in 2003. She is a Professor with the Department of Mechanical Engineering, Dalhousie University, Halifax, NS, Canada. She held the post-doctoral fellow positions with the Laboratoire d'Automatique de Grenoble, CNRS, Grenoble, France and the Department of Electrical and Computer Engineering, University of Alberta, Edmonton, Canada, respectively. Her research interests include robust nonlinear control, cyber-physical systems, intelligent transportation systems, haptics, and collaborative multiple robotic systems. She is a Fellow of the Canadian Academy of Engineering (CAE), a Fellow of the Engineering Institute of Canada (EIC), a Fellow of ASME, a Fellow of CSME, a Senior Member of IEEE, and a registered Professional Engineer in Nova Scotia, Canada.
\end{IEEEbiography}

\vspace{-1cm}
\begin{IEEEbiography}[{\includegraphics[width=1in,height=1.25in,clip,keepaspectratio]{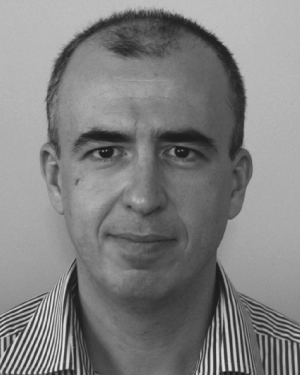}}]{Onur Toker} (Member, IEEE) received the B.S. degrees in electrical and electronics engineering, mathematics, and physics from Bogazici University, Istanbul, Turkey, and the M.S. degrees in electrical engineering and mathematics and the Ph.D. degree in electrical engineering from The Ohio State University, Columbus, OH, USA. He was a Graduate Researcher with the Technical University of Eindhoven, Eindhoven, The Netherlands, and subsequently with the University of California, Riverside, Riverside, CA, USA. He served as a faculty member at King Fahd University of Petroleum and Minerals (KFUPM), while also holding various consulting positions. He subsequently worked with Quanergy, a LiDAR technology company, Sunnyvale, CA, USA, and DAQRI, an augmented reality company, Sunnyvale, CA, USA, and Pasadena, CA, USA. He is currently a Professor of Electrical and Computer Engineering with Florida Polytechnic University, Lakeland, FL, USA. His research interests include FPGA-based hardware acceleration; millimeter-wave (mmWave) radar and communication systems; mmWave software-defined radios; and autonomous robotic systems. He has authored more than 100 technical publications and has served as the principal investigator (PI) or co-principal investigator (co-PI) on several funded research projects.
\end{IEEEbiography}

\end{document}